\documentclass[journal,twoside,web]{ieeecolor}

\usepackage{generic}
\usepackage{cite}
\usepackage{amsmath,amssymb,amsfonts}
\usepackage{algorithmic}
\usepackage{graphicx}
\usepackage{algorithm,algorithmic}
\usepackage{hyperref}
\hypersetup{hidelinks=true}
\usepackage{textcomp}
\usepackage{svg}
\usepackage{multirow}
\usepackage{gensymb}
\usepackage{url}
\usepackage{hyperref}
\usepackage{adjustbox}
\usepackage{graphbox}
\usepackage{xcolor,amsmath}
\usepackage{fontawesome}
\usepackage{mathtools}  
\usepackage{xfrac}  
\usepackage{tikz}

\def\BibTeX{{\rm B\kern-.05em{\sc i\kern-.025em b}\kern-.08em
    T\kern-.1667em\lower.7ex\hbox{E}\kern-.125emX}}
\definecolor{brickred}{HTML}{f03b20}

\definecolor{rubinered}{HTML}{CE0058}

\definecolor{bleudefrance}{rgb}{0.19, 0.55, 0.91}

\definecolor{mediumblue}{rgb}{0.0, 0.0, 0.8}

\definecolor{violet}{HTML}{6a51a3}

\usepackage[switch]{lineno}
\newboolean{highlighted}
\setboolean{highlighted}{false} % true: highlighted | false: unhighlighted
\ifthenelse{\boolean{highlighted}}{
\newcommand{\revision}[1]{\textcolor{blue}{#1}} % First pass approved by R1
}{
\newcommand{\revision}[1]{\textcolor{black}{#1}}
 
}
\begin{document}

\title{
\revision{
A Feasibility Study on Indoor Localization and Multi-person Tracking Using Sparsely Distributed Camera Network with Edge Computing}
}

\author{Hyeokhyen Kwon$^*$, 
Chaitra Hegde$^*$, 
Yashar Kiarashi,
Venkata Siva Krishna Madala,
Ratan Singh,
ArjunSinh Nakum,
Robert Tweedy,
Leandro Miletto Tonetto,
Craig M. Zimring,
Matthew Doiron,
Amy D. Rodriguez,
Allan I. Levey,
and Gari D. Clifford, \IEEEmembership{Fellow, IEEE}
\thanks{
$^*$ Equal contributions.
}
\thanks{
% This paragraph of the first footnote will contain the date on 
% which you submitted your paper for review. It will also contain support 
% information, including sponsor and financial support acknowledgment. 
% Dr. Allan I. Levey (Department of Neurology, School of Medicine, Emory University) has raised funds and provided the study space and environment (Emory Cognitive Empowerment Program at Goizueta Alzheimer's Disease Research Center). This work is also funded by Cox Foundation.
The Cognitive Empowerment Program is supported by a generous investment from the James M. Cox Foundation and Cox Enterprises, Inc., in support of Emory’s Brain Health Center and Georgia Institute of Technology.
}
\thanks{
% The next few paragraphs should contain 
% the authors' current affiliations, including current address and e-mail. 
Hyeokhyen Kwon, Yashar Kiarashi, and Robert Tweedy are with the Department of Biomedical Informatics, School of Medicine, Emory University, Atlanta, GA 30322 USA (email: hyeokhyen.kwon@dbmi.emory.edu; yash@dbmi.emory.edu; robert.tweedy@emory.edu).}
\thanks{
Chaitra Hegde, Venkata Siva Krishna Madala, Ratan Singh, and ArjunSinh Nakum are with the School of Electrical and Computer Engineering, Georgia Institute of Technology, Atlanta, GA 30332 USA (email: chegde@gatech.edu; vmadala3@gatech.edu; rsingh388@gatech.edu; arjun5inh@gatech.edu).}
\thanks{
Leandro Miletto Tonetto is with the School of Industrial Design, College of Design, Georgia Institute of Technology, Atlanta, GA 30332 USA (email: ltonetto3@gatech.edu).}
\thanks{
Craig M. Zimring is with the School of Architecture, College of Design, Georgia Institute of Technology, Atlanta, GA 30332 USA (email: craig.zimring@design.gatech.edu).}
\thanks{
Matthew Doiron, Amy D. Rodriguez, and Allan I. Levey are with the Department of Neurology, School of Medicine, Emory University, Atlanta, GA 30322 USA (email: matthew.james.doiron@emory.edu; amy.rodriguez@emory.edu; alevey@emory.edu).}
\thanks{
Gari D. Clifford is with the Department of Biomedical Informatics, Emory University School of Medicine, Atlanta, GA 30322 USA, and the Department of Biomedical Engineering, Georgia Institute of Technology and Emory University, Atlanta, GA 30322, USA (email: gari@gatech.edu).}
}

\maketitle

\begin{abstract}
\revision{
Camera-based activity monitoring systems are becoming an attractive solution for smart building applications with the advances in computer vision and edge computing technologies.
In this paper, we present a feasibility study and systematic analysis of a camera-based indoor localization and multi-person tracking system implemented on edge computing devices within a large indoor space. To this end, we deployed an end-to-end edge computing pipeline that utilizes multiple cameras to achieve localization, body orientation estimation and tracking of multiple individuals within a large therapeutic space spanning $1700m^2$, all while maintaining a strong focus on preserving privacy. Our pipeline consists of 39 edge computing camera systems equipped with Tensor Processing Units (TPUs) placed in the indoor space's ceiling. To ensure the privacy of individuals, a real-time multi-person pose estimation algorithm runs on the TPU of the computing camera system. This algorithm extracts poses and bounding boxes, which are utilized for indoor localization, body orientation estimation, and multi-person tracking \footnote{https://github.com/cliffordlab/EdgeCameraTracking}. 
Our pipeline demonstrated an average localization error of 1.41 meters, a multiple-object tracking accuracy score of 88.6\%, and a mean absolute body orientation error of 29\degree. These results shows that localization and tracking of individuals in a large indoor space is feasible even with the privacy constrains. 
}

\end{abstract}


\section{Introduction}

%-----------------------------------------
% [] Importance of vision-based localization and tracking
% [] Limitation of Previous work
% [] Contribution of our work
% [] Major motivation of the proposed work
% major contribution and results of the proposed work
% broader impact of the proposed work
%-----------------------------------------

Indoor localization and tracking is an active research area to support various applications, such as smart hospitals~\cite{haque2017towards,yang2020role}, smart offices~\cite{lopez2017human}, or customer analysis in shopping mall~\cite{dogan2019analyzing}. 
A wide range of sensors have been used to simultaneously detect the whereabouts and movement of multiple people in indoor spaces over the past two decades, including radio-frequency identification (RFID)~\cite{vuong2014automated}, infrared (IR)~\cite{cheol2017using}, WiFi~\cite{van2016performance}, and Bluetooth~\cite{yoo2018real}.
Recent advances in multi-person pose estimation~\cite{dai2022survey,dos2021monocular} and tracking have led to the use of cameras to monitor indoor activities~\cite{tsai2016vision,cosma2019camloc}, and
vision-based approaches have proven to be very effective in capturing accurate and detailed movements continuously over time.
\revision{
However, most studies have only examined a few cameras in small rooms and have not studied a system that integrates localization, tracking, and body orientation estimation techniques in a large indoor space required to understand fine-grained activities comprehensively ~\cite{luo2018computer,xue2018vision,haque2017towards}. Advancements in edge computing now allow for low-cost, powerful devices to monitor activity in existing buildings~\cite{jin2022survey,suresha2022edge,barthelemy2019edge}. 
}

\revision{
In this work, we present a feasibility study and systematic analysis of a camera-based indoor localization and tracking system to monitor multiple people across a wide indoor space using an edge computing framework. To carry out this study, we deployed a multi-camera edge computing pipeline designed to estimate locations, track individuals, and ascertain their body orientations in a vast therapeutic indoor space with a strong emphasis on safeguarding privacy. 
}
\revision{
We deployed 39 cameras sparsely positioned throughout an large indoor space spanning $1700m^2$ which contains diverse functional areas, such as a gym, library, and kitchen. These cameras were connected to edge computing units, specifically Raspberry Pi v4 B devices. 
}
\revision{
We adopted a real-time processing approach using Google Coral Tensor Processing Units (TPUs) connected to the edge computing units. These TPUs automatically process raw color frames obtained from the cameras to detect 2D poses (as described in~\cite{papandreou2018personlab}). This information is then used to estimate the positions and orientations of people within the indoor space, and the original raw frames are discarded. The estimated positions and orientations from all the cameras are combined to track the movement of individuals.
Using our camera infrastructure, we aim to provide a comprehensive analysis of the camera installation and its relations with the performance of our system in various activity scenarios.
We expect this work to provide practical guidelines for practitioners when installing a multi-view camera-based activity monitoring system using edge computing frameworks in existing facilities.
}

\section{Method \& Materials}

\begin{figure}[t]
    \centering
    \includegraphics[width=.9\linewidth]{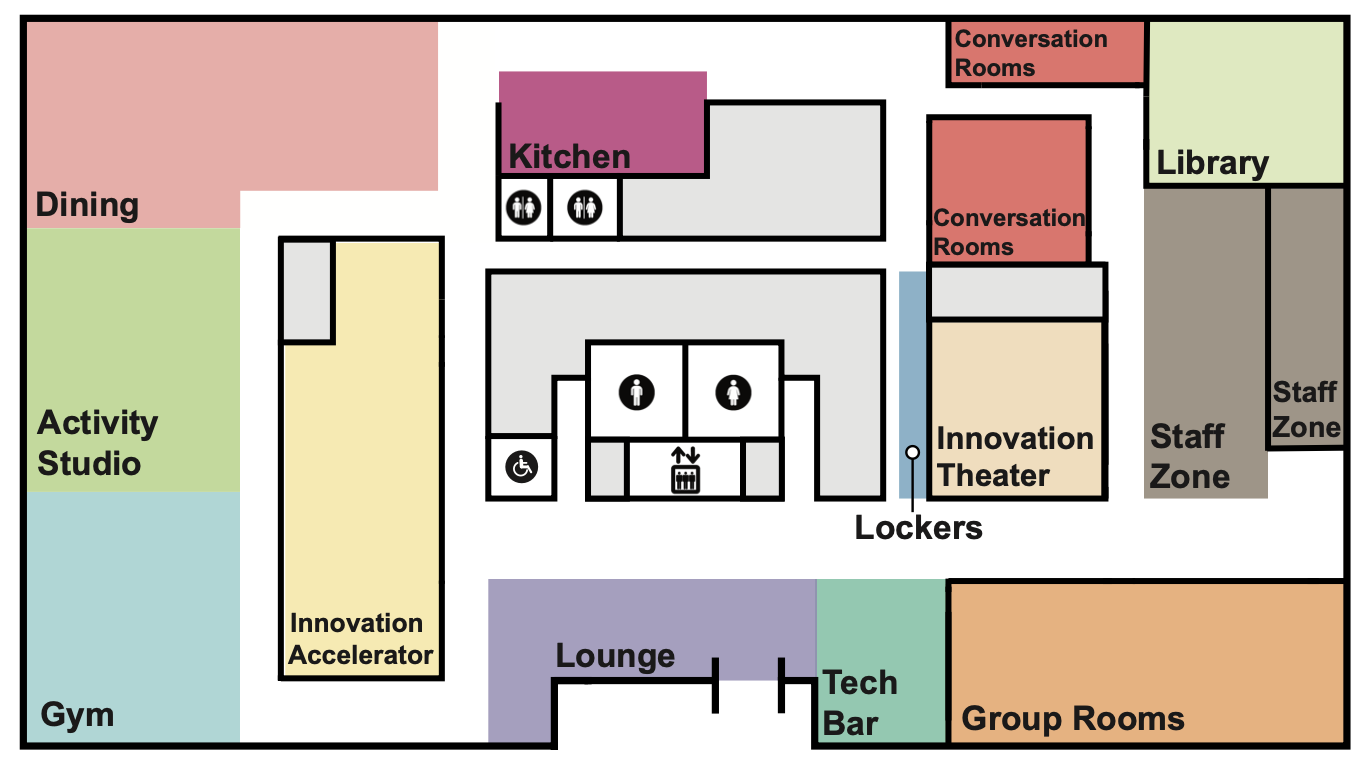}
    \vspace{-0.1in}
    \caption{
    \revision{
    Space usage in the study space. 
    }
    % There is a variety of functional spaces for people to conduct daily activities, including a library, kitchen, gym, and so on. 
    % Through the use of sparsely distributed multiple cameras and low-cost edge computers, we propose a system that can continuously and passively monitor the detailed movements of multiple people in space with minimum modification of existing infrastructure.    
    }
    \label{fig:ep6_func}
\end{figure}

% In this work, multiple edge computing-based camera systems are integrated into a distributed network and are installed with edge, fog, and cloud computing environments to locate and track multiple people with their body orientations in a built environment.
% This section provides a detailed account of our study site, the computing environment of our system, the techniques employed to continuously monitor people's movements in a given space, and an evaluation of the systems.

\subsection{Study Site}

% \revision{
% The location of our study is a 1700 square meter therapeutic indoor environment.
% }
Our site was designed to promote the involvement of older adults in a range of therapeutic activities, including physical exercise and memory training. It also provides opportunities for social interaction, such as shared meals and comfortable lounge areas.
The study site consists of diverse functional regions as shown in \autoref{fig:ep6_func}. The Activity Area (A) encompasses both the Activity Studio and Gym, promoting physical exercise. The Kitchen (B) covers the upper corridor between the dining area's right boundary and the conversation room's left boundary, encouraging independent cooking activities.
The left corridor (C) runs alongside a room called the Innovation Accelerator, while the Right Corridor (E) is situated next to the Lockers, to the left of the Innovation Theater. The lounge (D) extends from the elevator area to the front door, and the Staff area (F) comprises the section on the building's right side, where staff offices are located.

% This study site is specifically designed to help individuals with MCI to lead better lives. It is a one-of-a-kind project designed based on a series of goals: promoting social interaction, environmental exploration, spatial flexibility, safety, independent learning of everyday tasks (e.g., cooking), contact with nature, exercise, and cognitive stimulation in individuals diagnosed with MCI. The study spaces is adjustable and equipped with technologies like tunable lighting and controllable zone-based sound systems to experiment with achieving these goals. Designing healthcare facilities for individuals living with MCI poses challenges for architects as it requires unique design considerations that differ from those required for cognitively healthy older adults and dementia patients~\cite{challenge_arch_2020}. 
% While people with MCI also experience cognitive decline, their condition is less debilitating than that of dementia patients~\cite{Gauthier2006_classic}.

\subsection{Distributed Camera Network using Edge Computing}

\begin{figure}[t]
    \centering
    \begin{adjustbox}{width=1.0\columnwidth,center}
        \begin{tabular}{c}
            \includegraphics[align=c,width=1\linewidth]{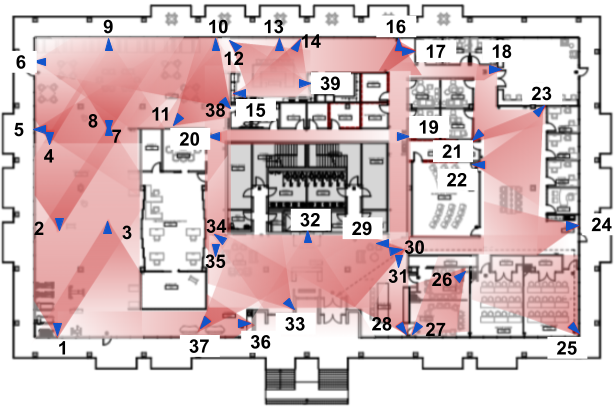} \\
            (a) \\
            \begin{tabular}{c c}
                \includegraphics[align=c,height=.4\linewidth]{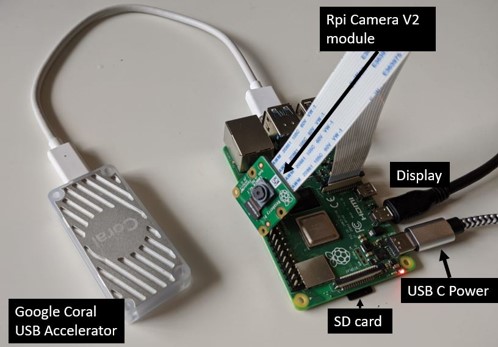} &
                \includegraphics[align=c,height=.4\linewidth]{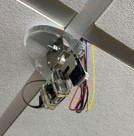} \\
                (b) & (c) 
            \end{tabular} 
        \end{tabular}
    \end{adjustbox}
    \vspace{-0.1in}
    \caption{Cameras installation in the study site.
    (a) In total, 39 camera sensors (Sony IMX219 8-megapixel sensor; \textcolor{blue}{BLUE}) are installed with the field of views (\textcolor{red}{RED}) covering the entire indoor space. %\hyeok{Update with the new figure.}
    (b) Raspberry Pi 4 is installed with Google Coral TPU USB Accelerator to run deep learning-based 2D pose estimation models in real time.
    (c) Example of camera installation in the ceiling.
    Raspberry Pi 4 is hidden in the ceiling and connected to the nearest network and power sources.
    }
    \label{fig:camera_config}
\end{figure}

% \begin{figure}[t]
%     \centering
%     \begin{adjustbox}{width=1.0\columnwidth,center}
%         \begin{tabular}{c c c}
%             \includegraphics[align=c,height=.4\linewidth]{figure/camera_config/camera_configuration.png} &
%             \includegraphics[align=c,height=.2\linewidth]{figure/camera_config/camera_ceiling.jpg} & 
%             \includegraphics[align=c,height=.2\linewidth]{figure/camera_config/camera_with_rpi.jpg} \\
%             (a) & (b) & (c) 
%         \end{tabular}
%     \end{adjustbox}
%     \vspace{-0.1in}
%     \caption{Cameras installation in the study site.
%     (a) In total, 38 camera sensors (Sony IMX219 8-megapixel sensor) are installed with the facing viewpoint toward the darker blue direction.
%     (b) Example of camera installation in the ceiling.
%     Raspberry Pi 4 is hidden in the ceiling and connected to the nearest network and power sources.
%     (c) Raspberry Pi 4 is installed with Google Coral USB Accelerator to run deep learning-based 2D pose estimation models in real-time.}
%     \label{fig:camera_config}
% \end{figure}

% Edge computing-based camera devices are installed in the ceiling of the study site and integrated with a server infrastructure. This system is described in this section. 

\subsubsection{Camera Configuration}

\revision{
\autoref{fig:camera_config}a shows the locations (\textcolor{blue}{BLUE}) of 39 low-cost ($<\$500$) edge computing and camera devices (\autoref{fig:camera_config}b) and their viewpoints (\textcolor{red}{RED}).
These cameras effectively provide coverage over the entire study site where subjects have consented to have their activities monitored. The cameras along with the edge devices were placed in the ceiling of the indoor environment away from the} \revision{direct line of sight of visitors to prevent them from feeling overly self-conscious, as shown in \autoref{fig:camera_config}c. These edge devices were connected to the closest available power and network sources in the ceiling through USB-C power cable and ethernet cable for data transfer. 
Depending on the availability of power and network sources, some locations, such as kitchen or staff areas, are covered by multiple cameras, and other locations, such as long corridors, are covered by a single or two cameras. 
The cameras installed above visitors' eyesight make it easier for visitors to comply with the proposed instrumentation of cameras for activity monitoring. 
}

\begin{figure*}[t]
    \centering
    \includegraphics[width=.75\linewidth]{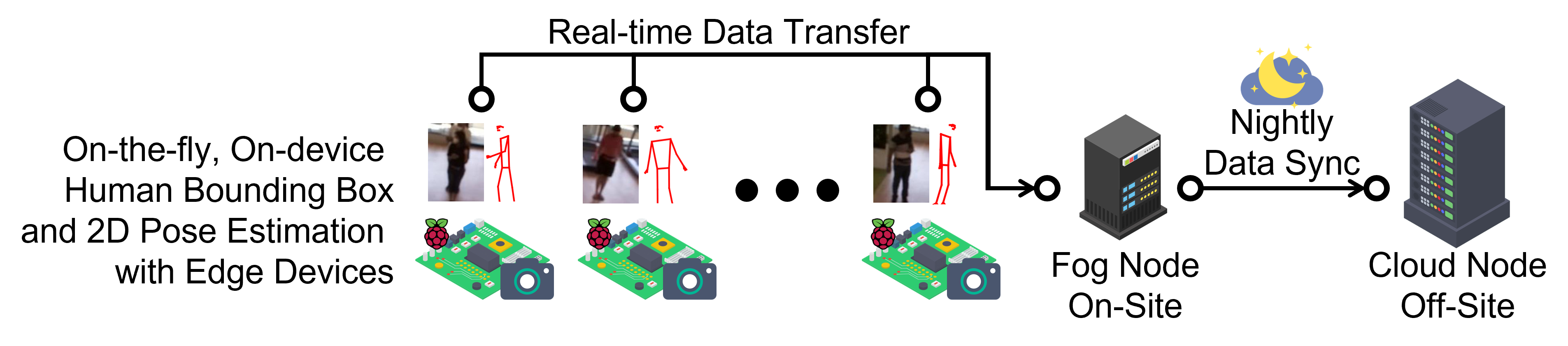}
    \vspace{-0.1in}
    \caption{
    \revision{
    Computing network architecture and data management in our study. 
    }
    % 39 edge devices estimate 2D poses and human bounding boxes on-the-fly without storing raw image frames on devices to preserve the privacy of monitored subjects.
    % The 2D pose and bounding box estimations are transferred to the on-site privacy-preserving fog computing node in real time.
    % The bounding box images are deleted as soon as they are processed to extract human body orientation.
    % These data collected each day in the fog node is synced nightly to the HIPAA-compliant cloud computing node for permanent storage and further analysis.
    }
    \label{fig:compute_config}
\end{figure*}

\begin{figure*}[t]
    \centering
    \includegraphics[width=0.95\linewidth]{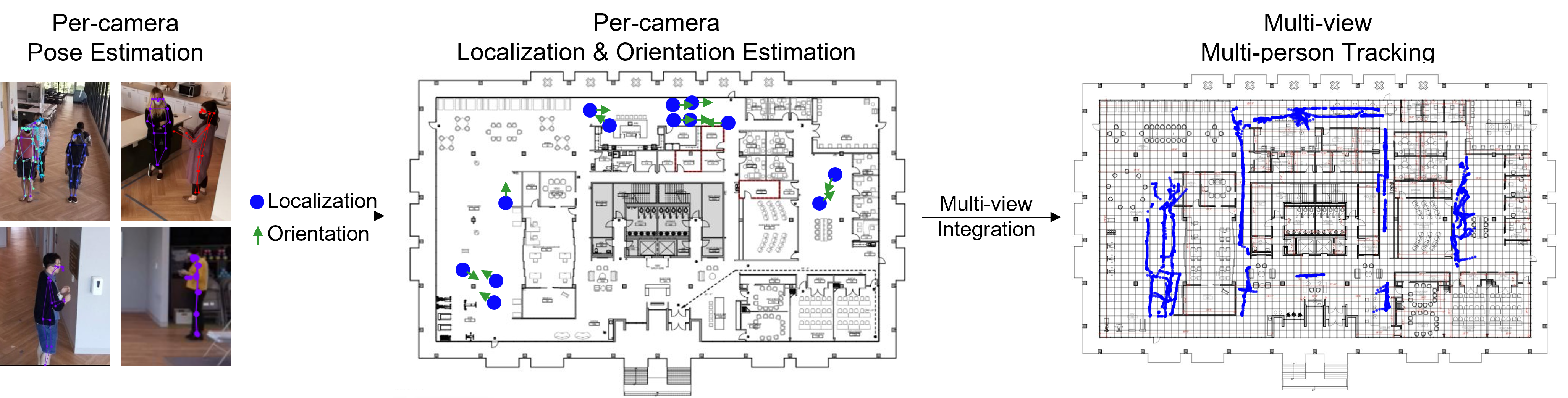}
    \vspace{-0.1in}
    \caption{Overall diagram of the proposed activity monitoring system.
    From each camera, the location and orientation of multi-person poses are projected into the study space's 2D coordinates.
    While visitors spend time in the study space, samples from multiple views are integrated and tracked.
    % The proposed method demonstrated localization error of less than 1 meter compared to the ground-truth location from our benchmark dataset.
    }
    \label{fig:overall_method}
\end{figure*}

\subsubsection{Secure Computing Network Architecture}
The complete computing setup comprises four key components: the off-site cloud server, the on-site fog server, the edge computing network infrastructure, and the array of 39 edge computing camera devices. Here, on-site and off-site servers refer to servers that are physically present in the study site and servers places outside of the study site respectively. This configuration is shown in \autoref{fig:compute_config}.

\revision{
Real-time estimation of 2D poses, composed of 17 keypoints following the MS COCO format~\cite{lin2014microsoft}, and bounding boxes occurs at a frame rate of 1Hz using the TPU on the edge computing device. Bounding box images are temporarily retained on the on-site fog computing node for up to one hour to facilitate body orientation estimation. They are then discarded. The raw frames are immediately discarded from the edge computing device after the 2D poses are extracted and do not get sent to the on-site fog server. This ensures privacy of individuals in the study site and prevents the capture, storage,} \revision{or transmission of sensitive or identifiable information, such as faces or nudity. The 2D poses and body orientations of a single day, specifically from 8 AM to 7 PM, are stored on the fog server and this data is routinely synchronized with the off-site cloud server during the night, typically between 12 AM and 7 AM. The duration of the nightly synchronization can change depending on the volume of data collected during the day. It's worth noting that the fog server complies with the Health Insurance Portability and Accountability Act (HIPAA) regulations. 
}

\subsection{Monitoring Movement in Built-in Environment}

\subsubsection{Overall Pipeline}
The overall approach for multi-view multi-person tracking using the above mentioned computing and camera infrastructure is shown in \autoref{fig:overall_method}.
Using the 2D poses and human bounding boxes, each person's location and body orientation is estimated and projected on a 2D coordinate system representing the study space.
Multi-view integration then resolves duplicate detections across cameras from overlapping views by combining samples from 39 cameras.
Lastly, a multi-person tracking algorithm~\cite{li2010multiple} is used to track people from the combined location and orientation samples.

\subsubsection{Preprocessing 2D poses}
\label{sec:preproc_poses}

The 2D poses estimated by the edge computing systems undergo a preprocessing step to eliminate noisy data. We identified two primary issues in the poses detected by the pose detection model~\cite{papandreou2018personlab}: non-human poses and temporal inconsistencies.
The model occasionally misinterprets patterns on the walls and floor as human poses due to its overhead camera perspective. These non-human poses remain static over time and show minimal pose changes. Consequently, we identify and remove these false positives (non-human samples) by examining subsequent frames for near-zero pose changes.
The second issue relates to the lack of temporal consistency in 2D poses because the model estimates poses frame by frame. Some poses exhibit discrepancies, such as left and right sides being flipped or even missing entirely between frames (false negatives). To address this, we implement temporal smoothing using a Kalman filter-based approach to reduce noise and fill in missing pose detections~\cite{10.1115/1.3662552,chan2011implementation}.

\subsubsection{Indoor Localization}
\label{sec:indoor_localization}
Our indoor localization technique is designed to determine the positions of individuals within the study site using a 2D coordinate system that represents the layout of the site, which we call the study site coordinate system. This is achieved by utilizing the 2D poses detected from each camera.
In our approach, we define the location of a person within a frame as the midpoint between their left and right feet. Given that the floor in our study area is uniformly flat, all individuals detected in a frame are assumed to be on the same plane.
To enable this localization, we derive a perspective transformation matrix $M\in \mathbb{R}^{3\times3}$. This matrix serves to project the foot locations from the 2D camera space to the corresponding floor positions within the study site coordinate system~\cite{sun2019see,cosma2019camloc}. The process for obtaining this transformation matrix $M$ is carried out manually for each camera by aligning four corresponding points in both the 2D camera space and the study space coordinate system.

% \begin{equation}
%     \begin{bmatrix}
%         x \\ y \\ 1
%     \end{bmatrix}
%     = \begin{bmatrix}
%         m_{11} & m_{12} & m_{13} \\
%         m_{21} & m_{22} & m_{23} \\
%         m_{31} & m_{32} & m_{33}
%     \end{bmatrix}
%     \begin{bmatrix}
%         u \\ v \\ 1
%     \end{bmatrix}  
%     \label{eqn:perspective_mat}
% \end{equation}

% \noindent where $[x, y]$ is the foot location in study site coordinate and $[u, v]$ is the foot pixel location in the camera.
% From every 39 cameras, we learn $M$ by manually selecting 4 corresponding points in the 2D coordinate study site and the pixel locations in the field of view of a camera~\cite{abdel2015direct,shim2015mobile}.
% The derived $M$s for each camera are deployed to localize the feet locations of detected 2D poses in the study space coordinate. 
% unless the camera location or orientation is changed.

\subsubsection{Body Orientation Estimation}
\label{sec:body_ori_est}

\begin{figure*}[t]    
    \centering
    \includegraphics[width=.8\linewidth]{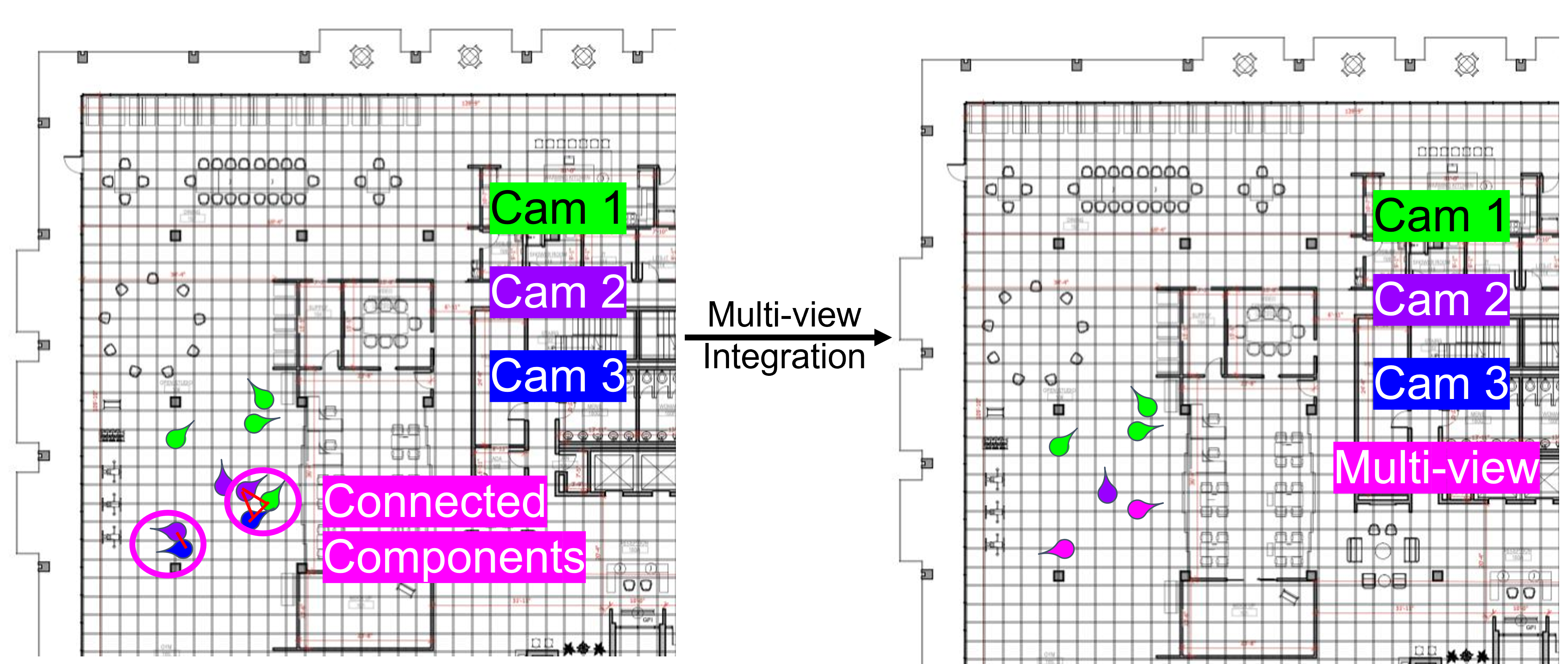}
    \caption{Multi-view integration of duplicated samples from multiple cameras having overlapping views.
    We first generate a graph, in which vertices are all samples from all cameras.
    Then the edges are created for samples from different cameras that are close together (within 1.5 meters) (Left, \textcolor{red}{red}).
    The connected components are then integrated as a single sample potentially representing the true location and orientation of the person in the study space (Right, \textcolor{magenta}{magenta}).
    }
    \label{fig:muti-view_integration}
\end{figure*}
Starting with the human bounding boxes obtained from 2D pose detection models, we extract the chest-facing direction of each individual within the transverse plane. This orientation is subsequently translated into the study space coordinate system.
To estimate the chest-facing direction from a human bounding box image, we employ a deep learning-based method for body orientation estimation~\cite{wu2020mebow}. Here, we define a human bounding box as a tightly enclosed square region that encompasses a detected 2D pose from the edge computing systems.
The resultant estimate of the chest orientation is represented as a 2D chest vector on the transverse plane. This 2D chest vector, initially defined in camera space, is then transformed into a 2D orientation in the study site coordinate system, denoted as $\Vec{v}^t_{world} = R_{camera \rightarrow world}\Vec{v}^t_{world}$. The rotation matrices, $R_{camera \rightarrow world}$, are manually determined for each camera based on the direction in which the camera is facing within the study site, with north serving as the reference orientation.

\subsubsection{Multi-view Integration}
To eliminate duplicated samples that may occur due to overlapping camera fields-of-view, we adopt a method that involves detecting people across multiple cameras and consolidating these detections into a single sample, potentially representing the person's actual location, as illustrated in \autoref{fig:muti-view_integration}.
To identify duplicate detections, we construct a graph, denoted as $G=(V,E)$. In this graph, the vertices, represented as $V=\{v_{ci}=(x,y)|c=1,\cdots,39, i=1,\cdots\}$, correspond to the locations of samples from all 39 cameras. The edges, denoted as $E=\{v_{ci}v_{c'j} |\; c \neq c' \}$, signify the connections between samples from different cameras that are sufficiently close in distance.
We assume that duplicates originate from the same person captured by different cameras if their Euclidean distance is within $\| v_{ci} - v_{c'j} \|_2 \leq 1.5 m$. This threshold takes into account the potential errors in our localization technique. While variations may occur due to lens distortions, which can lead to discrepancies in the perspective transformations, our empirical findings suggest that a margin of 1.5 meters is adequate~\cite{zhang2000flexible}.

Once the graph $G$ is established, duplicate samples are identified as connected components (CC) within the graph, following principles from graph theory~\cite{john1995first}. The integration of duplicate sample locations considers the distance of cameras from the detected sample. More precisely, during the integration of duplicated samples, a penalty is applied when the distance between the estimated person's location and the camera is substantial. This approach acknowledges the strong correlation between localization errors and the distance between the camera and the target person in our analysis. This is represented by the equation:

\begin{equation}
    L = \sum_{CC} \frac{1}{d_{ci}} v_{ci}
\end{equation}
 
\noindent Here, $L$ denotes the integrated sample location, and $d_{ci} = \| l_c - v_{ci} \|^2_2$ signifies the Euclidean distance between the sample, $v_{ci}$, and the camera, $l_c$.

Similarly, the orientation of the duplicated samples is integrated for the vertices within the connected components (CC), taking camera distances into account. This multi-view integration process results in unique locations, denoted as $L=[l_x, l_y]$, and orientation vectors, $o=[o_x, o_y]$ for each person at each time step.

\subsubsection{Multi-person Tracking}

\begin{figure*}[t]
    \centering
    \begin{adjustbox}{width=1.0\linewidth,center}
        \begin{tabular}{c c c}
            \includegraphics[align=c,width=.5\linewidth]{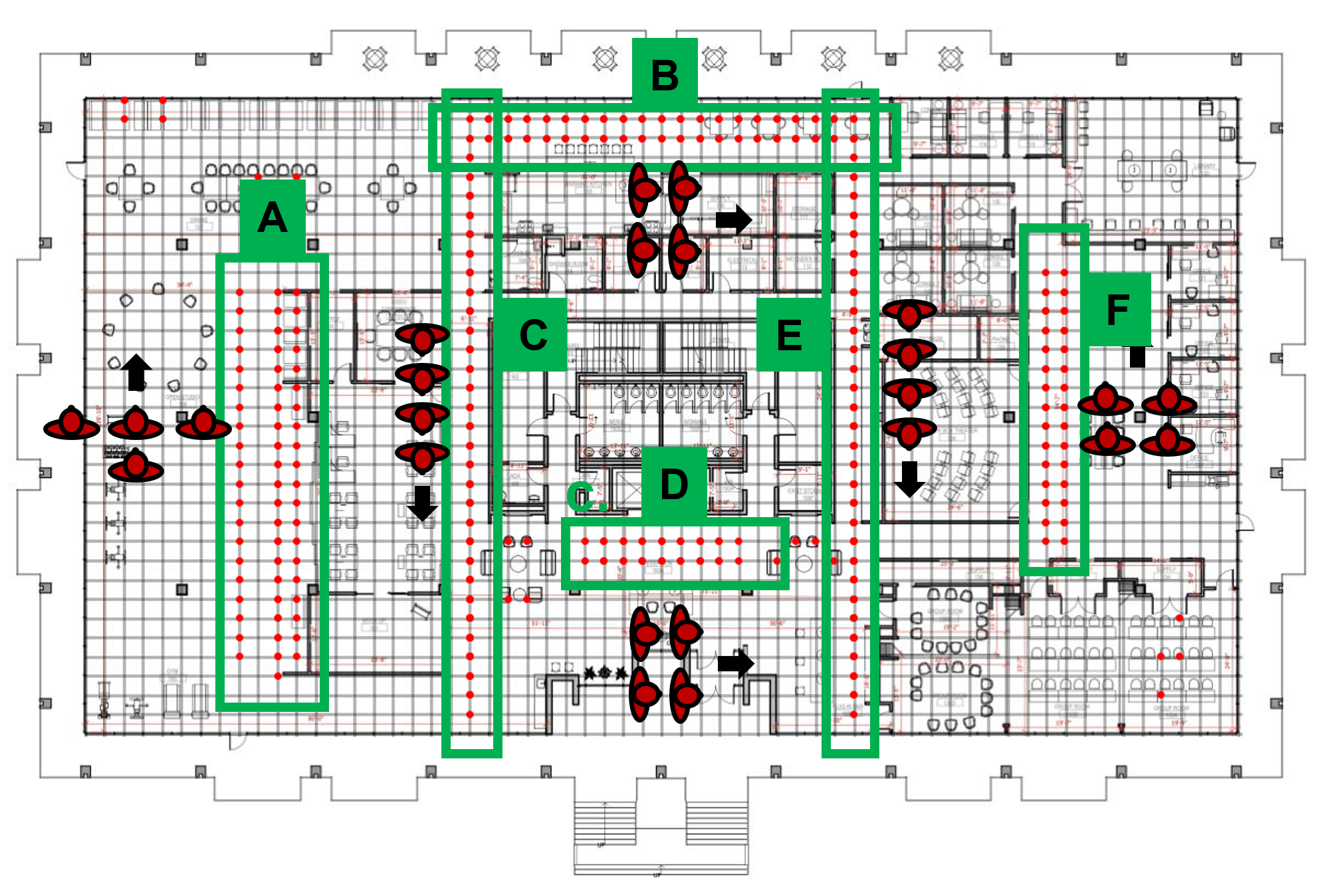} &
            \includegraphics[align=c,width=.15\linewidth]{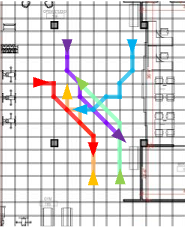} &
            \includegraphics[align=c,width=.35\linewidth]{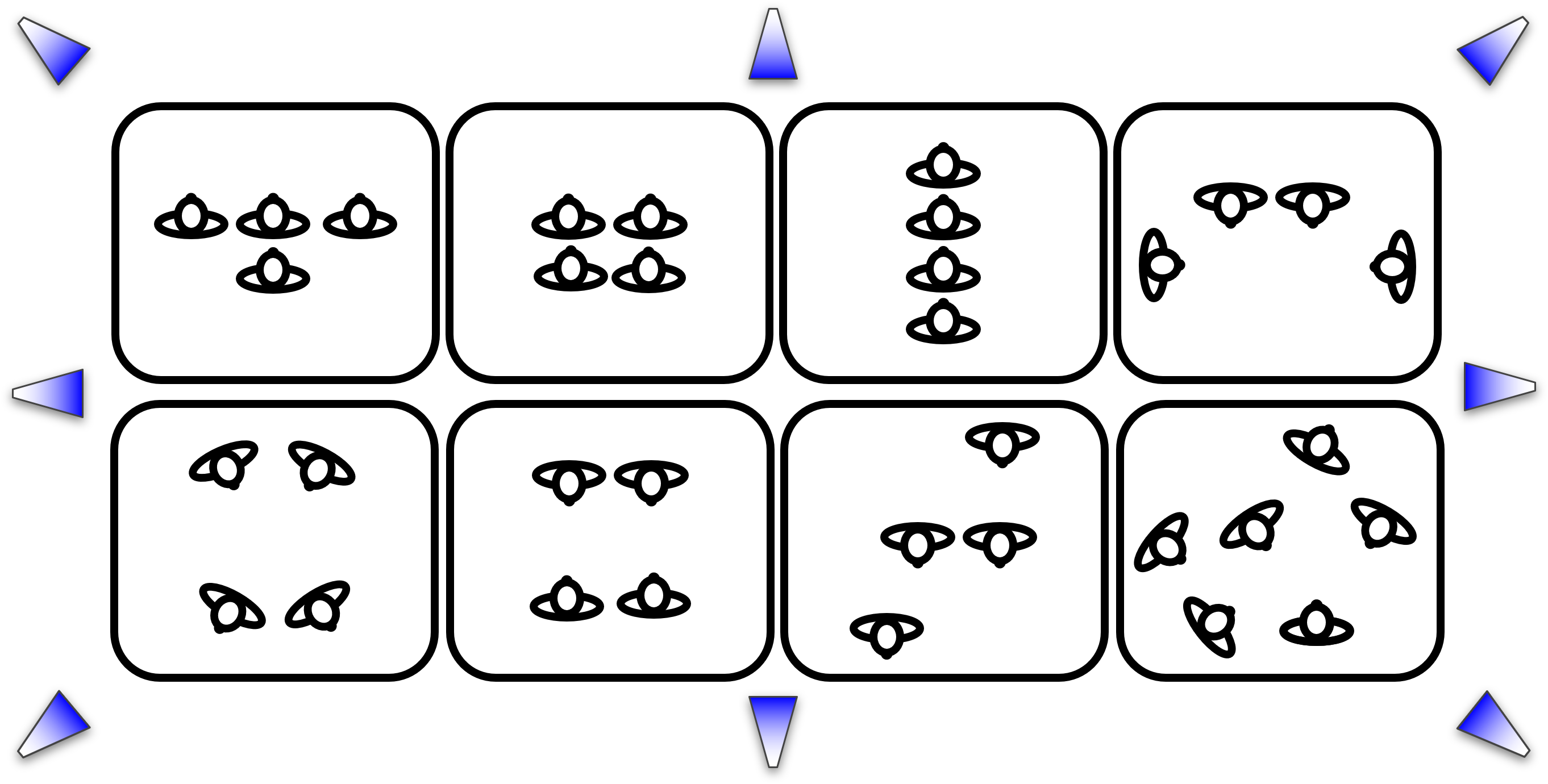} \\
            (a) & (b) & (c)
        \end{tabular}
    \end{adjustbox}
    \vspace{-0.1in}
    \caption{
    \revision{
    Benchmark data collection.
    We selected the group formations in order to simulate realistic scenarios during the study.
    (a) Four participants (\textcolor{red}{Red}) were walking with different formations while collecting the benchmark dataset.
    Our team has covered nearly all areas of the study site where visitors spend a majority of their time.
    (b) Five participants showing more complex movements with occlusions. 
    % We simulated a scenario in the activity area where multiple people walk toward and past each other with occlusions. 
    This was collected in Activity Area, shown in (a) A, where most physical activity occurs in our study site.
    Each person is color-coded with different colors. 
    (c) Overall examples of group formations taken by participants and camera viewpoints while collecting the benchmark data.
    % The distributed cameras across the study site captured movements from multiple viewpoints (\textcolor{blue}{Blue}).
    % Our analysis shows that having a dense installation of cameras that can capture people from a close distance ($\sim 15m$) with a frontal view ($<30\degree$ horizontal angle) is crucial.    
    }
    }
    \label{fig:data_collection}
\end{figure*}
We implement Kalman filter-based tracking for the locations denoted as $L=[l_x, l_y]$ and the orientation vector represented by $O=[o_x, o_y]$. This approach assumes that people's facing orientation closely follows their walking direction~\cite{kuhn1955hungarian} and can be expressed as follows:

\begin{equation}
    O_t= O_{t-1} + w\cdot \dot{L}_{t-1} + dt\cdot \dot{O}_{t-1}
\end{equation}
 
\noindent Here, the parameter $w=0.1$ acts as a weighting factor for the influence of the movement direction (or velocity) on the current orientation, and $dt$ denotes the time interval between frames, which is set at 1 second. 
To further enhance tracking performance, we apply Rauch–Tung–Striebel (RTS) smoothing to completed tracks, which takes into account temporal dynamics across the entire trajectory of a person.

\subsection{Benchmark Data Collection}\label{sec:benchmark_data_collection}
In order to methodically assess the effectiveness of our system for localization, body orientation estimation, and tracking, we created an annotated dataset. This dataset features four to five individuals moving and interacting within the study site.
The benchmark dataset encompasses scenarios where people are 1) walking together, 2) walking past each other, or 3) conversing in place while adjusting their facing orientations.
\revision{
We marked the entire area with 1-meter intervals to provide ground-truth reference points. 
}
The subjects were instructed to follow predetermined group formations and designated walking paths.
As the individuals moved within the space, an observer recorded the time and location of each subject at one-second intervals. Additionally, the observer manually annotated the body orientation of the subjects, using the north direction (as indicated by the upper side of \autoref{fig:ep6_func}) as a reference point.
Below, we outline the three distinct datasets we generated to capture various activities within the space.

\subsubsection{4 People Walking around Entire Space}
In the first benchmark dataset, we had four individuals moving throughout the study site. To ensure that our dataset encompasses realistic use cases for visitors in the study site, we carefully selected locations and paths as depicted in \autoref{fig:data_collection}. These areas were of particular interest for tracking behaviors within the study space (\autoref{fig:data_collection}a). These locations included spaces frequently used by visitors for activities like dining, kitchen usage, physical exercises, relaxation, and staff zones (\autoref{fig:ep6_func}), where participants had given their consent to be recorded.

For annotation purposes, we utilized the closest 1-meter markers as the ground-truth reference points for the subjects while they were walking. This level of detail was sufficient to account for variations in foot size and stride length among the participants.
As the subjects walked, we introduced changes in their group formation to simulate realistic movement patterns within the space (\autoref{fig:data_collection}c). In wider corridors, we arranged two or three people to walk side by side, while in narrower corridors, multiple subjects proceeded in straight lines. This approach allowed us to assess the impact of occlusions, where subjects might not be visible from certain camera viewpoints.

\subsubsection{5 People Walking Past Each other}
As depicted in \autoref{fig:data_collection}b, we recreated a scenario in the activity area where multiple individuals walked towards one another and passed by each other, a situation that often results in occlusions. Such occlusions are common during activities like dancing or group exercises.

\subsubsection{5 people Changing Orientations in Place}

\begin{figure*}[t]
    \centering
    \begin{tabular}{c}
        Session 1 (Group size: 2 \& 3) \\
        \includegraphics[align=c, width=.7\linewidth]{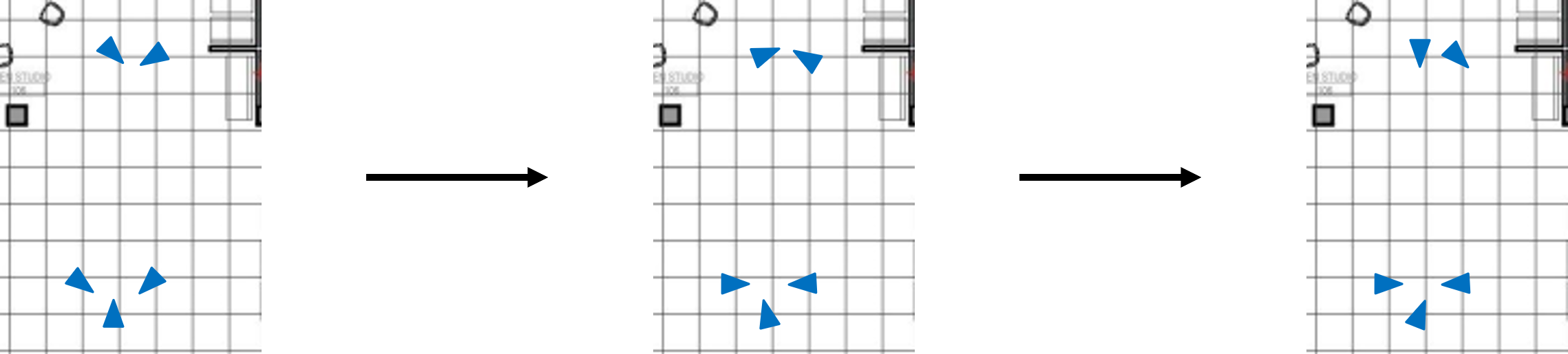} 
        \vspace{0.05in}\\
        Session 2 (Group size: 4) \\
        \includegraphics[align=c, width=.7\linewidth]{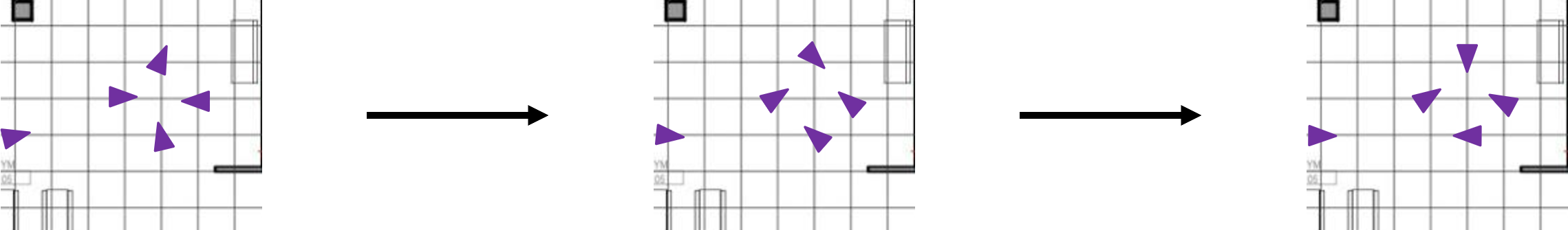} 
        \vspace{0.05in}\\
        Session 3 (Group size: 5) \\
        \includegraphics[align=c, width=.7\linewidth]{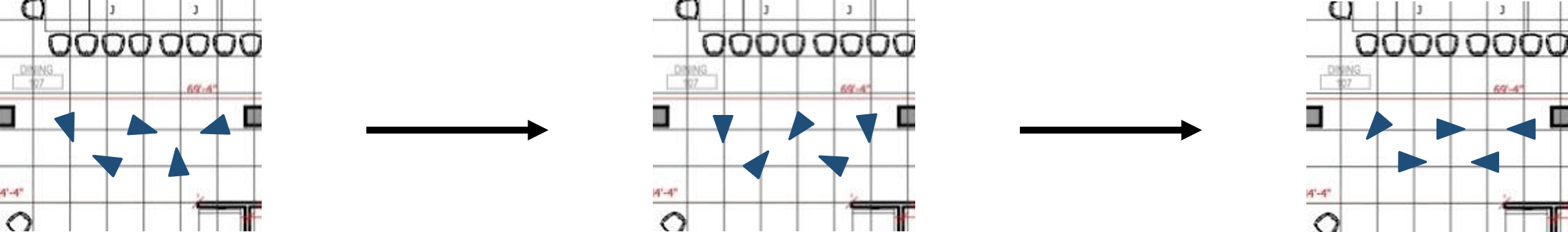} \\
    \end{tabular}
    \vspace{-0.0in}
   \caption{
   Data collection when people make orientation changes while standing.
   We capture three sessions from different locations in the Activity Area, with different sizes of social groups.
   The first session had two social groups with two and three members, the second session had one social group with four members, and the third session had one social group with five members.
   Participants were instructed to change their facing directions during conversations.
   }
    \label{fig:ori_in_place}
\end{figure*}
\begin{figure*}[t]
    \centering
    \begin{adjustbox}{width=0.95\textwidth,center}
    \begin{tabular}{c c c c}
        Distance between &
        Facing angle &
        Horizontal  &
        Vertical  \\
        camera and target (m) &
        away from camera &
        angle/Field-of-View &
        angle/Field-of-View \\
        \includegraphics[align=c, height=0.2\linewidth]{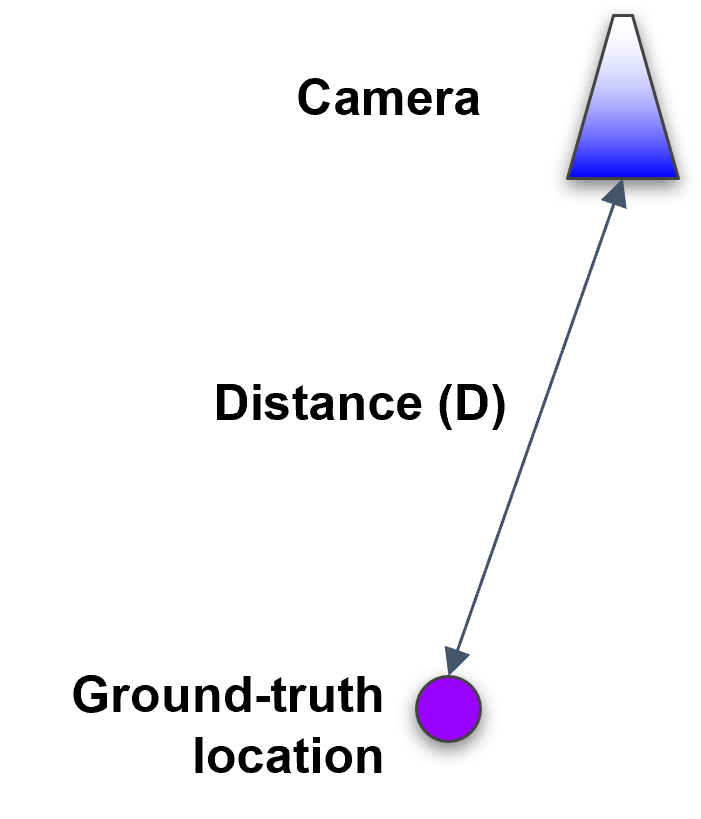} &
        \includegraphics[align=c, height=0.2\linewidth]{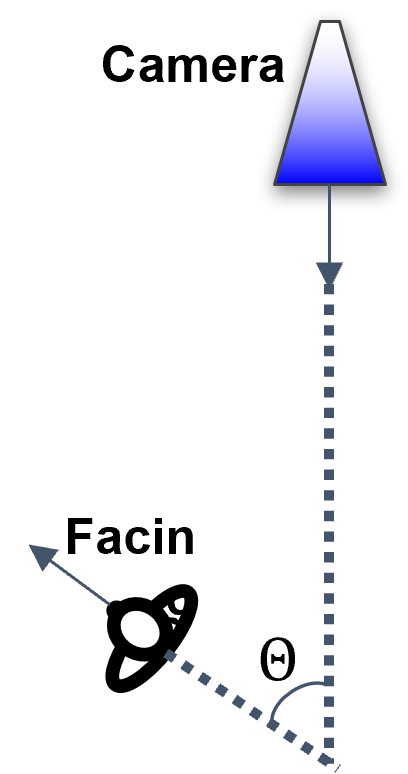} &
        \includegraphics[align=c, height=0.2\linewidth]{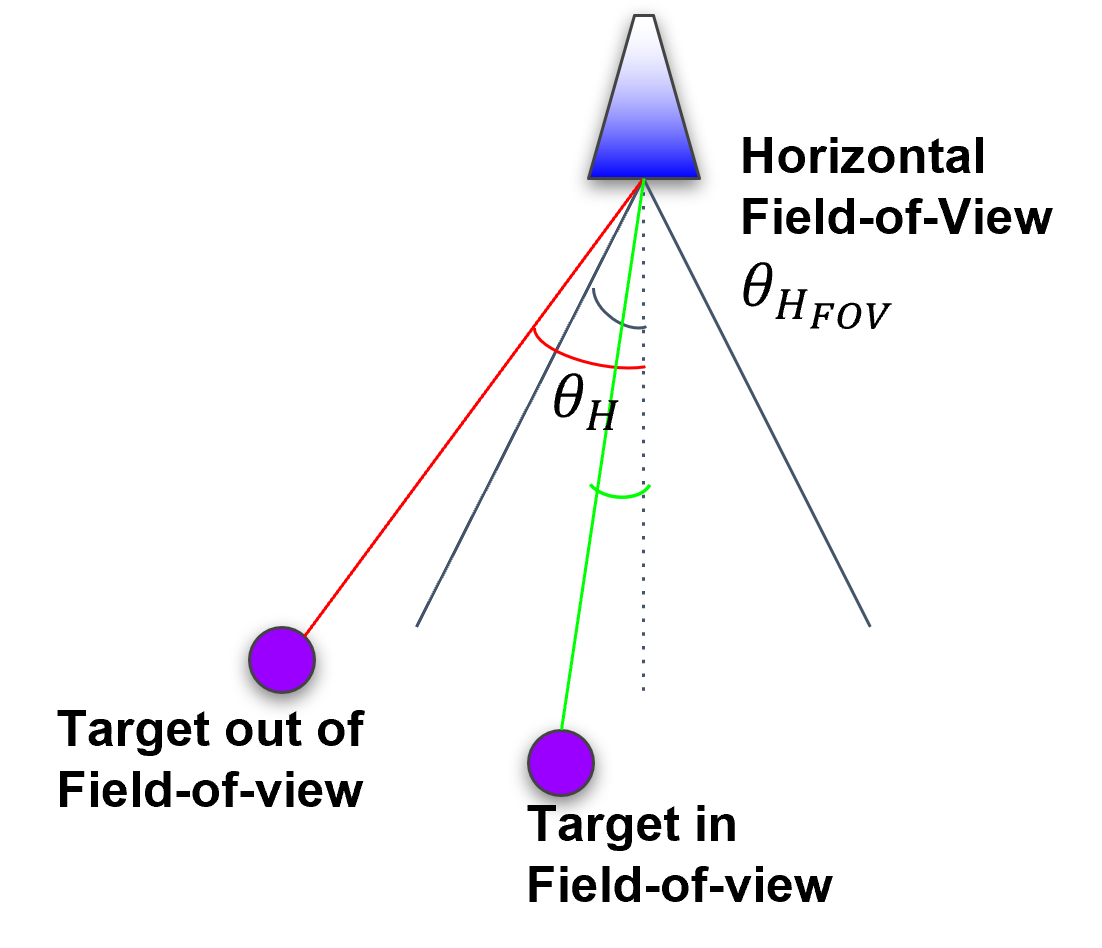} &
        \includegraphics[align=c, height=0.2\linewidth]{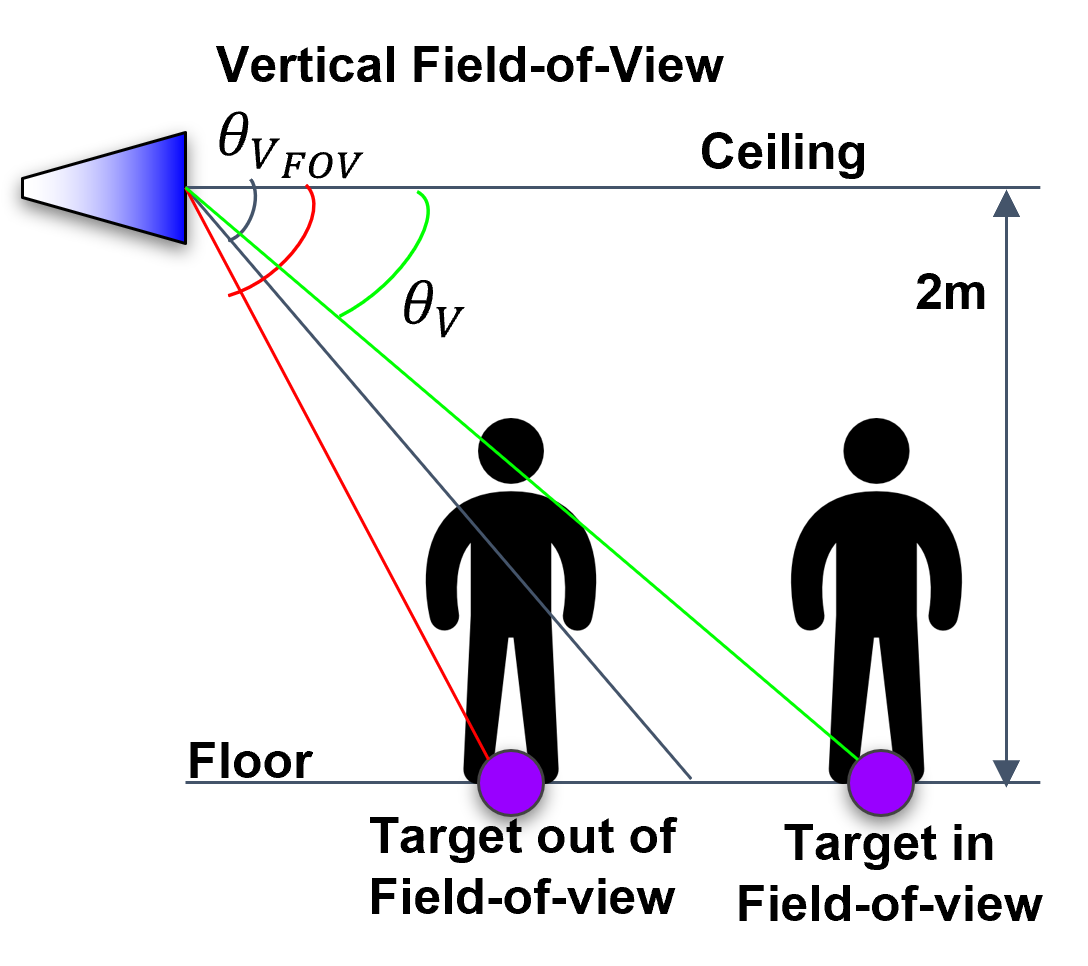} \\
        (a) & (b) & (c) & (d)
    \end{tabular}
    \end{adjustbox}
    \vspace{-0.1in}
    \caption{Analysing the impact of camera installation with respect to four aspects.    
    Those include (a) the distance between the camera and the person in the space, (b) a person's facing direction away from the camera, (c) the horizontal angle of the captured person from the camera center, and (d) vertical angle of the captured person's feet location from camera center.
    The horizontal angle (c; $\theta\degree_H$) and vertical angle (d; $\theta\degree_V$;) are normalized by horizontal ($\theta\degree_{H_{FOV}}$) and vertical ($\theta\degree_{V_{FOV}}$) field-of-view angles, respectively, referring to the camera datasheet (Sony IMX219 8-megapixel sensor).}
    \label{fig:cam_install_protocol}
\end{figure*}
During group conversations, individuals often remain seated or stand stationary, adjusting their facing directions to interact. In this dataset, we focused on situations where people made subtle changes in their facing orientation while staying in one spot, as shown in \autoref{fig:ori_in_place}.
We conducted this dataset in three sessions, starting with a smaller social group of two participants and gradually increasing it to five. In this context, a social group refers to people engaging in active communication within a distance of less than 2 meters, taking into account pandemic-related social distancing guidelines. Participants followed instructions to gradually modify their facing direction as part of the data collection process.

\subsection{Evaluating Multi-person Tracking and Localization}
We assess the performance of our activity monitoring system by examining its localization, tracking, and body orientation estimation capabilities, following established evaluation methods~\cite{bernardin2006multiple,wu2020mebow}.
Since there isn't a singular metric for evaluating multi-person tracking methods, we employ a set of metrics to comprehensively evaluate the proposed system~\cite{dendorfer2020mot20}. In the evaluation, metrics denoted by $\uparrow$ indicate better performance when higher, while those marked with $\downarrow$ signify better performance when lower.

\begin{itemize}
    \item MOTA $\uparrow$ (Multi-Object Tracking Accuracy) combines three kinds of error sources, which are false positives, missed targets, and identity switches.    
    \item IDF1 $\uparrow$ (ID F1 score) is the ratio of correctly identified detections over the average number of ground-truth and computed detections.
    \item MT $\uparrow$ (Mostly Tracked targets) measures the ratio of ground-truth trajectories that are covered by a track hypothesis for at least 80\% of their respective life span.
    \item ML $\downarrow$ (Mostly Lost targets) measures the ratio of ground-truth trajectories that are covered by a track hypothesis for at most 20\% of their respective life span.
    \item FPR $\downarrow$ (False Positives Rate) measures the average number of false alarms per frame.
    \item FNR $\downarrow$ (False Negatives Rate) measures the average number of missed targets per frame.
    \item Rcll $\uparrow$ (Recall) measures the ratio of correct detections to the total number of ground-truth locations.
    \item Prcn $\uparrow$ (Precision) measures the number of detected objects over the sum of detected and false positives.
    \item IDSR $\downarrow$ (ID Switch Ratio) measures the number of ID switches (IDS) over recall.
    \item Frag $\downarrow$  (Fragmentation) measures the total number of times a trajectory is fragmented, which are interruptions during tracking.
\end{itemize}

In particular, we assess the system's performance using the MOTA metric, defined as follows:
\begin{equation}
    MOTA=1-\frac{\sum_t FN_t + FP_t + IDS_t}{\sum_t GT_t}
\end{equation}

\noindent Here, $FN_t$, $FP_t$, $IDS_t$, and $GT_t$ represent false negatives (missed detections), false positives, ID switches, and ground-truths at time $t$, respectively.

An essential evaluation that precedes tracking is detection. Here, we define a ground-truth target at time $t$ as "detected" if the estimated sample location falls within a 1.5-meter Euclidean distance. Otherwise, the target is considered "missed." The choice of a 1.5-meter threshold is based on the observation that people typically walk at speeds below 1.42 m/s in everyday situations~\cite{browning2006effects,mohler2007visual,levine1999pace}.
To showcase the accuracy of the estimated sample locations, we also report the multiple object tracking precision, denoted as MOTP:

\begin{equation}
    MOTP=\frac{\sum_{i,t} d^i_t}{\sum_t c_t}
\end{equation}
 
\noindent Here, $d^i_t$ represents the Euclidean distance between the $i$th detected ground-truth sample and the estimated sample, and $c_t$ is the number of detections at time $t$. In simpler terms, MOTP provides the average Euclidean distance error for the detected samples, indicating the system's ability to estimate the precise location of target samples, independent of tracking trajectories or ID switches.

% \subsubsection{Multi-person Tracking Baseline}

% Our proposed tracker is compared with a Hungarian matching-based tracker~\cite{kuhn1955hungarian}, which only considers frame-by-frame matching between 2D pose detections, in order to illustrate the importance of modeling temporal dynamics of 2D movements.
% Based on the results of multiview multi-person localization at time $t$ and $t+1$, the Hungarian tracker generates a bipartite graph with nodes representing samples from two consecutive timesteps and edges representing Euclidean distances between samples. 
% Since our tracking rate is 1 Hz, we remove any edge distance larger than 1.5 meters in the created bipartite graph, as people normally walk 1.42 m/s in everyday activities~\cite{browning2006effects,mohler2007visual,levine1999pace}.
% Then, the Hungarian method finds the matchings (or edges) that have a minimum sum of Euclidean distances.
% Potentially missing samples (false negatives) in a frame $t$ are filled in by linearly interpolating the matched detection from frame $t-1$ and $t+1$.
% The Hungarian matching is applied until no matching is found any longer over time.

\subsection{Evaluating Body Orientation Estimation}
Consistent with previous definitions of body orientation found in earlier works~\cite{wu2020mebow,yu2019continuous,andriluka2010monocular}, we represent human orientation, denoted as $\theta$, within the range of $[0\degree, 360\degree)$. This angle corresponds to the direction in which the chest is facing in the study site coordinate system. The north (top) direction serves as the reference point, with clockwise rotation applied.
For the assessment of body orientation, we utilize two metrics as also employed in previous studies~\cite{wu2020mebow}:
\begin{itemize}
    \item Mean Absolute Error (MAE) of Angular Distance: This metric quantifies the average absolute angular difference between predicted and ground-truth orientations.
    \item Accuracy-$X\degree$: Accuracy-$X\degree$ is defined as the proportion of predictions falling within a deviation of $X\degree$ compared to the ground-truth orientation. By considering different values of $X$ within the range of $[0\degree, 180\degree]$, this metric provides a fine-grained measure of model accuracy. Specifically, we evaluate Accuracy-$X\degree$ for a set of $X$ values including $X\in\{5\degree, 15\degree, 22.5\degree, 30\degree, 45\degree, 90\degree\}$~\cite{wu2020mebow}.
\end{itemize}

\subsection{Impact of Camera Installation}

% In our study site, the camera installation is determined primarily by the power and network sources reachable using extension cables, with minimum modification to the existing building structure, resulting in 39 cameras unequally installed in the space.
\revision{
We conducted a quantitative analysis to delve deeper into the influence of camera placement on four distinct factors that impact the performance of our system, as illustrated in \autoref{fig:cam_install_protocol}. In particular, we aim to establish the correlation between the subjects' positions and facing orientations as captured within the video frames of each camera, and how these aspects affect the accuracy of both localization and body orientation estimation. This analytical approach sheds light on the error budget arising from per-camera indoor localization (Section~\ref{sec:indoor_localization}) and body orientation estimation (Section~\ref{sec:body_ori_est}), ultimately contributing to our understanding of the overall system performance.
}

\subsubsection{Distance Between Camera and Person}
For all cameras, we assessed the distance between individuals captured in relation to the camera's center, as depicted in \autoref{fig:cam_install_protocol}a. This distance measurement is instrumental in evaluating the impact on localization and orientation errors.

\subsubsection{Facing Angle with respect to Camera Viewpoint}

\begin{figure*}[t]
    \centering
    \begin{tabular}{c c}
        \includegraphics[align=c, height=.3\linewidth]{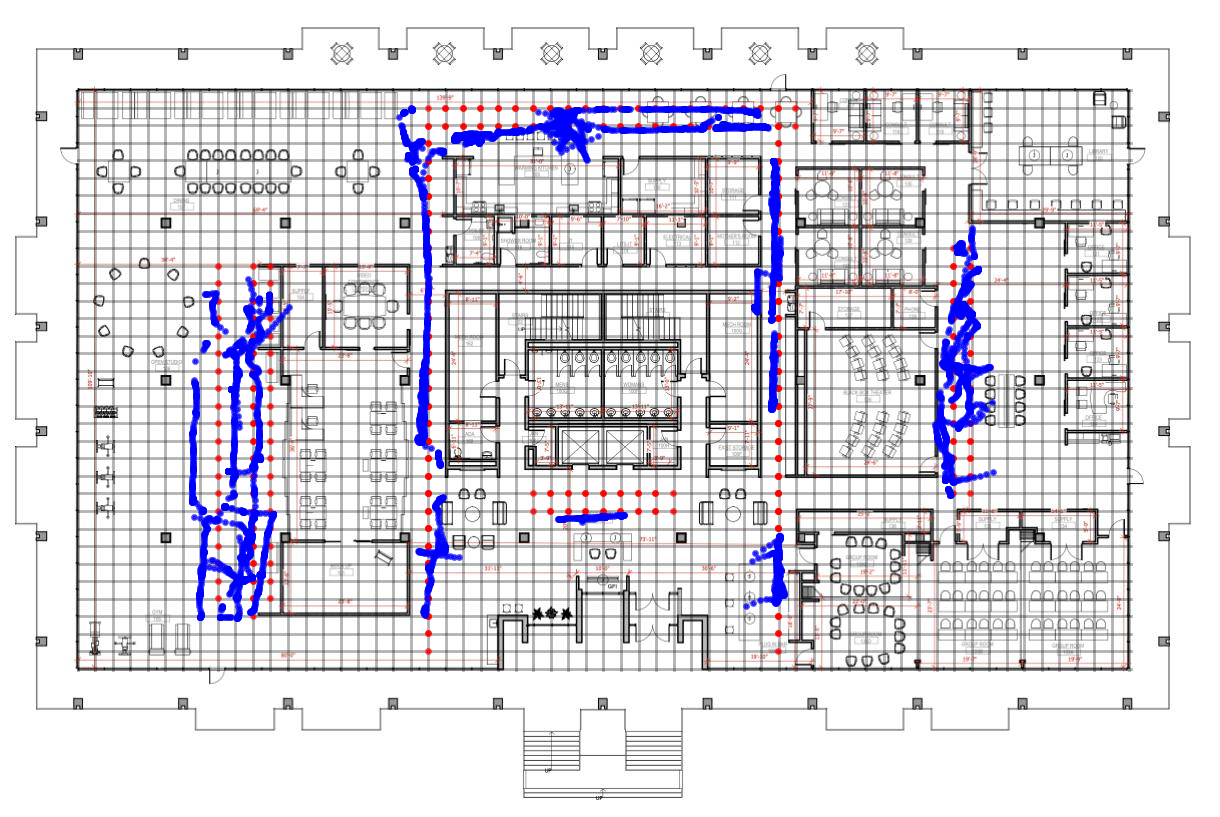} &
        \includegraphics[align=c, height=.27\linewidth]{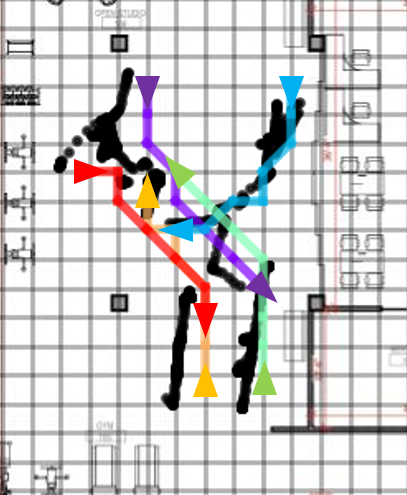} \\
        (a) & (b)        
    \end{tabular}
    \vspace{-0.0in}
   \caption{
   \revision{
   Qualitative results of multi-person tracking.
   (a) Four people walking around the study site following \autoref{fig:data_collection}a.
   The \textcolor{red}{REDs} are ground truth and \textcolor{blue}{BLUEs} are predicted location of people.
   % The estimated locations of people are within 2 meters from ground truths.
   % For the areas with only a few (one or two) cameras, such as Lounge and Right Corridors, we can observe discontinuities in trajectories.
   % Due to the distance between the camera and the target person, the pose estimation model failed to detect the target.
   (b) Five people walking in a complex pattern. 
    % This dataset involves multiple people passing each other and making occlusions as they cross paths.   
    Ground-truth trajectories are shown in colored lines with arrows indicating walking directions, while predicted trajectories are shown in black lines.
    % Having four cameras capturing the Activity Area, our proposed model could closely track complex movements having 79\% MOTA score and 70\% IDF1 score.   
   }
   }
    \label{fig:mot_result}
\end{figure*}
\begin{table*}[t]
    \centering
    \caption{
    \revision{
    Multi-object localization and tracking performance at each region in the study site when four participants walked around.
    The area sizes for each section are shown with height and width in meters.
    \faCamera\;
    % \yash{did we use all the cameras for all the captured frame. Let's say we have 7 cameras in the activity area, do we have footage from all 7 cameras in our dataset? if not, probably we need to point this out in the text.}\hyeok{Yes. We have footage of course and we use all of them as discussed in the method section.}
    shows the number of cameras that were covering the movement over 85\% of trajectories.
    % Localization performance is measured by multiple object tracking precision (MOTP) and tracking performance was measured by multiple metrics to capture the detailed behavior of the proposed method.
    Each area is annotated referring to \autoref{fig:data_collection}.
    % Activity Area (A) includes both Activity Studio and Gym area, and Kitchen (B) includes the upper corridor between the right boundary of the dining area and the left boundary of the conversation room.
    % The Left corridor (C) is the narrow corridor right next to the Innovation Accelerator, and Right Corridor (E) is the narrow corridor left of the Innovation Theater.
    % Lounge (D) includes the area between the elevator and the reservation desk.
    % Staff area (F) is in the right region where staff offices are located.
    % Here we demonstrate the performance of both our baseline (Hungarian) and proposed (Kalman Filter) trackers. 
    \textit{Overall} performance shows average metrics across all areas combined.
    % Compared to Kalman Filter-based trackers, Hungarian-based trackers do not take into account temporal dynamics of detections over time and are significantly affected by ID switching and missing samples.
    The last row additionally shows the quantitative results for simulating complex movements from five people with occlusions in Activity Area.
    % \hyeok{Done:Update FP and FN to per frame average}
    }
    }
    \vspace{-0.0in}
    \begin{adjustbox}{width=\textwidth,center}
        \begin{tabular}{c c ||c|c||c|c|c|c|c|c|c|c|c|c|c}          
            &
            Area & 
            Size ($H\times W$) &
            \faCamera &
            MOTP $\downarrow$ &
            MOTA $\uparrow$ & 
            IDF1 $\uparrow$ & 
            MT $\uparrow$ & 
            ML $\downarrow$ & 
            FPR $\downarrow$ & 
            FNR $\downarrow$ & 
            Rcll $\uparrow$ & 
            Prcn $\uparrow$ & 
            IDSR $\downarrow$ & 
            Frag $\downarrow$ \\
            \hline\hline
            A &
            Activity Area & 
            $21m \times 5m$ & % Size
            7 & % \faCamera
            $1.38m$ & % MOTP $\downarrow$
            0.886 & % MOTA $\uparrow$  
            0.584 & % IDF1 $\uparrow$  
            4 & % MT $\uparrow$  
            0 & % ML $\downarrow$  
            0.17 & % FP $\downarrow$  94/541
            0.25 & % FN $\downarrow$  138/541
            0.936 & % Rcll $\uparrow$  
            0.956 & % Prcn $\uparrow$  
            9.61 & % IDSR $\downarrow$ 
            6 \\ % Frag $\downarrow$
            \hline
            B &
            Kitchen & 
            $3m\times 24m$ & % Size
            3 & % \faCamera
            $1.48m$ & % MOTP $\downarrow$
            0.671 & % MOTA $\uparrow$  
            0.427 & % IDF1 $\uparrow$  
            1 & % MT $\uparrow$  
            0 & % ML $\downarrow$  
            0.28 & % FP $\downarrow$  161/571
            1.01 & % FN $\downarrow$  576/571
            0.747 & % Rcll $\uparrow$  
            0.913 & % Prcn $\uparrow$  
            17.38 & % IDSR $\downarrow$ 
            11 \\ % Frag $\downarrow$
            \hline
            C &
            Left Corridor & 
            $33m\times 2m$ & % Size
            3 & % \faCamera
            $1.92m$ & % MOTP $\downarrow$
            0.545 & % MOTA $\uparrow$  
            0.435 & % IDF1 $\uparrow$  
            1 & % MT $\uparrow$  
            0 & % ML $\downarrow$  
            0.27 & % FP $\downarrow$  149/551
            1.53 & % FN $\downarrow$  845/551
            0.616 & % Rcll $\uparrow$  
            0.901 & % Prcn $\uparrow$  
            12.97 & % IDSR $\downarrow$ 
            4 \\ %Frag $\downarrow$
            \hline
            D &
            Lounge & 
            $3m \times 8m$ & % Size
            1 & % \faCamera
            $0.80m$ & % MOTP $\downarrow$
            0.334 & % MOTA $\uparrow$  
            0.501 & % IDF1 $\uparrow$  
            0 & % MT $\uparrow$  
            2 & % ML $\downarrow$  
            0 & % FP $\downarrow$  0/151
            2.66 & % FN $\downarrow$  402/151
            0.334 & % Rcll $\uparrow$  
            1.0 & % Prcn $\uparrow$  
            0.0 & % IDSR $\downarrow$ 
            0 \\ %Frag $\downarrow$
            \hline
            E &
            Right Corridor & 
            $33m\times 1.5m$ & % Size
            2 & % \faCamera
            $1.75m$ & % MOTP $\downarrow$
            0.602 & % MOTA $\uparrow$  
            0.468 & % IDF1 $\uparrow$  
            2 & % MT $\uparrow$  
            0 & % ML $\downarrow$  
            0.01 & % FP $\downarrow$  9/551
            1.56 & % FN $\downarrow$  859/551
            0.610 & % Rcll $\uparrow$  
            0.993 & % Prcn $\uparrow$  
            13.10 & % IDSR $\downarrow$ 
            9 \\ %Frag $\downarrow$
            \hline
            F &
            Staff Zone & 
            $14m\times 3m$ & % Size
            4 & % \faCamera
            $1.11m$ & % MOTP $\downarrow$
            0.734 & % MOTA $\uparrow$  
            0.659 & % IDF1 $\uparrow$  
            3 & % MT $\uparrow$  
            0 & % ML $\downarrow$  
            0.65 & % FP $\downarrow$  205/311
            0.40 & % FN $\downarrow$  125/311
            0.899 & % Rcll $\uparrow$  
            0.845 & % Prcn $\uparrow$  
            4.44 & % IDSR $\downarrow$ 
            1 \\ %Frag $\downarrow$
            \hline\hline
            &
            Overall & 
            - & % Size
            - & % \faCamera
            $\mathbf{1.41m}$ & % MOTP $\downarrow$
            \textbf{0.629} & % MOTA $\uparrow$  
            \textbf{0.512} & % IDF1 $\uparrow$  
            1.833 & % MT $\uparrow$  
            0.333 & % ML $\downarrow$  
            0.23 & % FP $\downarrow$  
            1.23 & % FN $\downarrow$  
            0.690 & % Rcll $\uparrow$  
            0.934 & % Prcn $\uparrow$  
            9.58 & % IDSR $\downarrow$ 
            5.16 \\ %Frag $\downarrow$
            \hline\hline
            &
            5 people & 
            $13m \times 6m$ & % Size
            4 & % \faCamera
            $1.44m$ & % MOTP $\downarrow$
            0.798 & % MOTA $\uparrow$  
            0.700 & % IDF1 $\uparrow$  
            5 & % MT $\uparrow$  
            0 & % ML $\downarrow$  
            0.65 & % FP $\downarrow$  219/336
            0.34 & % FN $\downarrow$  114/336
            0.932 & % Rcll $\uparrow$  
            0.877 & % Prcn $\uparrow$  
            6.43 & % IDSR $\downarrow$ 
            1 \\ %Frag $\downarrow$
        \end{tabular}
    \end{adjustbox}
    \label{tab:mot}
\end{table*}

We gathered data on the extent to which the facing directions of individuals deviate from the camera viewpoints, as visualized in \autoref{fig:cam_install_protocol}b. Subsequently, we performed a correlation analysis to assess the relationship between these deviations and the associated localization and orientation errors.

\subsubsection{Horizontal Displacement from Camera Center}
We studied the impact of the horizontal displacement of individuals from the camera center, as illustrated in \autoref{fig:cam_install_protocol}c. To quantify this horizontal displacement, we introduced a normalized horizontal angle, denoted as $\sfrac{\theta\degree_H}{\theta\degree_{H_{FOV}}}$, where $\theta\degree_H$ represents the horizontal angle of the person from the camera center, and $\theta\degree_{H_{FOV}}$ is the horizontal field-of-view.

When a person moves closer to the horizontal center of the camera, the normalized horizontal angle approaches zero. Conversely, it increases to a value of $1$ when the person reaches the edge of the frame. If the value exceeds $1$, it indicates that only a portion of the person's body is visible within the camera's field of view.
We explore how localization and orientation errors are influenced by the degree of horizontal displacement.

\subsubsection{Vertical Displacement from Camera Center}
We investigated the impact of vertical displacement of an individual from the camera center, as indicated in \autoref{fig:cam_install_protocol}d. To measure this vertical displacement, we introduced a normalized vertical angle denoted as $\sfrac{\theta\degree_V}{\theta\degree_{V_{FOV}}}$. Here, $\theta\degree_V$ represents the vertical angle at which the person is captured from the camera center, and $\theta\degree_{V_{FOV}}$ is the vertical field-of-view, which is obtained from the camera datasheet.
We specifically focused on the foot location as a reference point to calculate the normalized vertical angle. When a person is closer to the camera, the normalized vertical angle increases, and only the upper body is partially visible in the camera frame due to the limited vertical field of view. Conversely, as the person moves away from the camera, the normalized vertical angle decreases, allowing for the full body to be captured within the field of view. However, this comes at the cost of increased distance between the camera and the person.
We examined how changes in localization and orientation errors are influenced by the vertical displacement of the captured individual.

\section{Results}

% In this section, we demonstrate the performance of the proposed localization, tracking, and body orientation estimation methods using the metrics described earlier. 
% We also compare our methods with the baseline methods.

\subsection{Multi-person Tracking and Localization}

\subsubsection{4 People Walking around Entire Space}

\begin{table*}[t]
    \centering
    \caption{
    \revision{
    Body orientation estimation performance.
    % We measured the Mean Absolute Error (MAE) and Accuracy-$X\degree$.
    % The predictions are considered correct when the angular error is within $X\degree$ of the ground-truth orientation. 
    % Following Wu~\etal~\cite{wu2020mebow}, we used $X\in\{5\degree, 15\degree, 22.5\degree, 30\degree, 45\degree, 90\degree\}$.
    The first six rows show the body orientation estimation performance at each region while four people walk in the space, and the seventh row shows the overall performance considering the entire area.
    % While people walk in the study site, the model can closely estimate the body orientation with MAE of 28.97\degree and Acc.-45\degree of 78.9\% which shows that our model can distinguish eight (N, NE, E, SE, S, SW, W, NW) orientations with sufficient margin of errors.
    5 people moving with a more complex trajectory in Activity Area are shown in the eighth row.
    % 5 people moving with a more complex trajectory in Activity Area (8th row) resulted in an MAE of 48.33\degree.
    % Our model has been significantly confused by occlusion between people, as detected bounding boxes include multiple people with different orientations.
    % Yet, our model can identify equally spaced four (N, E, S, W) orientations with sufficient margin of error showing Acc.-90\degree of 81.4\%.
    The last four rows show the estimation performance when a person stands in space but rotates only their orientations and the first three of these four rows show the evaluation from three different sessions. The last row shows the aggregated results.
    % Our model found it more challenging to estimate when a person is in place.
    % In these cases, our model could classify equally spaced four (N, E, S, W) with tighter margin of error showing MAE of 63.4\degree and Acc.-90\degree of 69.3\%.
    % Due to our method's frame-by-frame approach, it was difficult to detect fine orientation changes between $5\degree\sim 10\degree$ across frames.
    % Generally, our model performed better when people were walking as the Kalman Filter updates the orientation information to incorporate moving directions as the user moves.
    }
    }    
    \vspace{-0.0in}
    \begin{adjustbox}{width=\textwidth,center}
        \begin{tabular}{c c||c|c||c||c|c|c|c|c|c}
            &
            Area & 
            Size ($H\times W$) &
            \faCamera &
             MAE-$\theta\degree$ &
             Acc.-$5\degree$ &
             Acc.-$15\degree$ &
             Acc.-$22.5\degree$ &
             Acc.-$30\degree$ &
             Acc.-$45\degree$ &
             Acc.-$90\degree$ \\
             \hline\hline
             A &
             Activity Area & 
            $21m \times 5m$ & % Size
            7 & % \faCamera
             25.91\degree & % MAE-$\theta\degree$ &
             0.319 & %Acc.-$5\degree$
             0.622 & %Acc.-$15\degree$
             0.723 & %Acc.-$22.5\degree$
             0.800 & %Acc.-$30\degree$
             0.855 & %Acc.-$45\degree$
             0.912 \\ %Acc.-$90\degree$ 
             \hline
             B &
             Kitchen & 
            $3m\times 24m$ & % Size
            3 & % \faCamera
             31.06\degree & % MAE-$\theta\degree$ &
             0.207 & %Acc.-$5\degree$
             0.489 & %Acc.-$15\degree$
             0.590 & %Acc.-$22.5\degree$
             0.650 & %Acc.-$30\degree$
             0.748 & %Acc.-$45\degree$
             0.916 \\ %Acc.-$90\degree$ 
             \hline
             C &
             Left Corridor & 
            $33m\times 2m$ & % Size
            3 & % \faCamera
             21.68\degree & % MAE-$\theta\degree$ &
             0.223 & %Acc.-$5\degree$
             0.578 & %Acc.-$15\degree$
             0.706 & %Acc.-$22.5\degree$
             0.793 & %Acc.-$30\degree$
             0.887 & %Acc.-$45\degree$
             0.955 \\ %Acc.-$90\degree$ 
             \hline
             D &
             Lounge & 
            $3m \times 8m$ & % Size
            1 & % \faCamera
             47.37\degree & % MAE-$\theta\degree$ &
             0.038 & %Acc.-$5\degree$
             0.137 & %Acc.-$15\degree$
             0.302 & %Acc.-$22.5\degree$
             0.423 & %Acc.-$30\degree$
             0.554 & %Acc.-$45\degree$
             0.846 \\ %Acc.-$90\degree$ 
             \hline
             E &
             Right Corridor & 
            $33m\times 1.5m$ & % Size
            2 & % \faCamera
             32.41\degree & % MAE-$\theta\degree$ &
             0.214 & %Acc.-$5\degree$
             0.479 & %Acc.-$15\degree$
             0.556 & %Acc.-$22.5\degree$
             0.602 & %Acc.-$30\degree$
             0.674 & %Acc.-$45\degree$
             0.946 \\ %Acc.-$90\degree$ 
             \hline
             F &
             Staff Zone & 
            $14m\times 3m$ & % Size
            4 & % \faCamera
             33.68\degree & % MAE-$\theta\degree$ &
             0.096 & %Acc.-$5\degree$
             0.324 & %Acc.-$15\degree$
             0.464 & %Acc.-$22.5\degree$
             0.589 & %Acc.-$30\degree$
             0.778 & %Acc.-$45\degree$
             0.942 \\ %Acc.-$90\degree$ 
             \hline\hline
             &
             Overall & 
            - & % Size
            - & % \faCamera
             \textbf{28.97\degree} & % MAE-$\theta\degree$ &
             0.220 & %Acc.-$5\degree$
             0.506 & %Acc.-$15\degree$
             0.615 & %Acc.-$22.5\degree$
             0.693 & %Acc.-$30\degree$
             0.789 & %Acc.-$45\degree$
             0.930 \\ %Acc.-$90\degree$ 
             \hline\hline
             &
             5 people & 
            $13m \times 6m$ & % Size
            4 & % \faCamera
             48.33\degree & % MAE-$\theta\degree$ &
             0.149 & %Acc.-$5\degree$
             0.328 & %Acc.-$15\degree$
             0.417 & %Acc.-$22.5\degree$
             0.512 & %Acc.-$30\degree$
             0.629 & %Acc.-$45\degree$
             0.814 \\ %Acc.-$90\degree$ 
             \hline\hline
             &
             In-place (Sess. 1) & 
            $11m \times 5m$ & % Size
            3 & % \faCamera
             73.73\degree & % MAE-$\theta\degree$ &
             0.087 & %Acc.-$5\degree$
             0.181 & %Acc.-$15\degree$
             0.255 & %Acc.-$22.5\degree$
             0.261 & %Acc.-$30\degree$
             0.322 & %Acc.-$45\degree$
             0.644 \\ %Acc.-$90\degree$ 
             \hline
             &
             In-place (Sess. 2) & 
            $4m \times 7m$ & % Size
            2 & % \faCamera
             65.28\degree & % MAE-$\theta\degree$ &
             0.078 & %Acc.-$5\degree$
             0.212 & %Acc.-$15\degree$
             0.236 & %Acc.-$22.5\degree$
             0.290 & %Acc.-$30\degree$
             0.375 & %Acc.-$45\degree$
             0.672 \\ %Acc.-$90\degree$ 
             \hline
             &
             In-place (Sess. 3) & 
            $3m \times 5m$ & % Size
            4 & % \faCamera
             50.47\degree & % MAE-$\theta\degree$ &
             0.111 & %Acc.-$5\degree$
             0.265 & %Acc.-$15\degree$
             0.454 & %Acc.-$22.5\degree$
             0.580 & %Acc.-$30\degree$
             0.692 & %Acc.-$45\degree$
             0.769 \\ %Acc.-$90\degree$ 
             \hline\hline
             &
             In-place (Overall) & 
            - & % Size
            - & % \faCamera
             63.40\degree & % MAE-$\theta\degree$ &
             0.091 & %Acc.-$5\degree$
             0.218 & %Acc.-$15\degree$
             0.310 & %Acc.-$22.5\degree$
             0.371 & %Acc.-$30\degree$
             0.457 & %Acc.-$45\degree$
             0.693 \\ %Acc.-$90\degree$ 
        \end{tabular}
    \end{adjustbox}
    \label{tab:orientation}
\end{table*}
In this section, we present the system's overall performance when four people walk together throughout the entire study site. The results of multi-person tracking and localization are depicted in \autoref{fig:mot_result}, and detailed in \autoref{tab:mot}, where each functional area is assessed individually.
It's worth noting that different locations within the study site have varying sizes of the captured area (as shown in the second column of \autoref{tab:mot}) and a different number of cameras (as indicated in the third column of \autoref{tab:mot}), all of which captured 85\% of the samples.
The overall MOTA score stands at 62.9\%, and the localization performance, measured by MOTP, is $1.41m$. These results indicate that our system achieved accurate localization and tracking in over 60\% of cases, taking into account false positives, false negatives, and identity switches.
A more detailed analysis of the impact of camera installation on localization accuracy is provided in Section~\ref{sec:cam_install}.

\subsubsection{5 People Walking Past Each other}

% \input{table/mot_traj}

% We evaluated multi-person tracking performance when multiple people were showing more complex movements.
% We simulated a scenario in the activity area where multiple people walk toward and past each other with occlusions.
% As mentioned earlier, Activity Area is the location where most physical activities are occurring in the study space, and also the region where our pipeline showed the highest accuracy due to denser camera installation.
Both qualitative and quantitative outcomes are presented in \autoref{fig:mot_result}b and in the last row of \autoref{tab:mot}, respectively. Despite the challenging scenarios, the system demonstrated robust performance when individuals were passing by each other.
The MOTA score, MOTP score, and IDF1 score reached 79.8\%, 1.44m, and 70\%, respectively, indicating strong performance with minimal ID switch errors in situations where people were crossing paths.

\subsection{Body Orientation Estimation}
The results of body orientation estimation are provided in \autoref{tab:orientation}. Our evaluations encompass scenarios involving four individuals walking across the entire space and five individuals passing by each other in the Activity Area. Additionally, we conducted assessments of body orientation for a scenario in which five individuals remained stationary while altering their orientations within different group formations.

\subsubsection{4 People Walking around Entire Space}
During the walking scenario with four participants, the overall mean absolute error (MAE) in body orientation estimation was 28.97$\degree$, and the accuracy within a 45$\degree$ margin reached 78.9\%, as indicated in the seventh row of \autoref{tab:orientation}. Further examination of the correlation between camera installation and body orientation estimation is presented in Section~\ref{sec:cam_install}.

\subsubsection{5 People Walking Past Each other}
In situations where individuals were walking closely past each other, our body-orientation estimation technique faced challenges due to the presence of multiple people with different orientations captured within the same human bounding box. As illustrated in the eighth row of \autoref{tab:orientation}, this posed difficulties, resulting in a high mean absolute error (MAE) of 48.33$\degree$ and a lower accuracy within a 45$\degree$ margin (62.9\%).

\subsubsection{5 people Changing Orientations in Place}
The final four rows in \autoref{tab:orientation} present the estimation performance when an individual remains stationary but rotates their orientation. In this scenario, the overall performance indicated an mean absolute error (MAE) of 63.40$\degree$.

\begin{figure*}[t]
    \centering
    \begin{adjustbox}{width=\textwidth,center}
    \begin{tabular}{c@{\hspace{1mm}} c c c c c}
        & % y label 
        & % y label
        Distance between &
        Facing angle &
        Horizontal  &
        Vertical  \\
        & % y label 
        & % y label
        camera and target (m) &
        away from camera &
        angle/Field-of-View &
        angle/Field-of-View \\
        \rotatebox[origin=c]{90}{Localization} &
        \rotatebox[origin=c]{90}{error (meter)} &
        \includegraphics[align=c, height=0.24\linewidth]{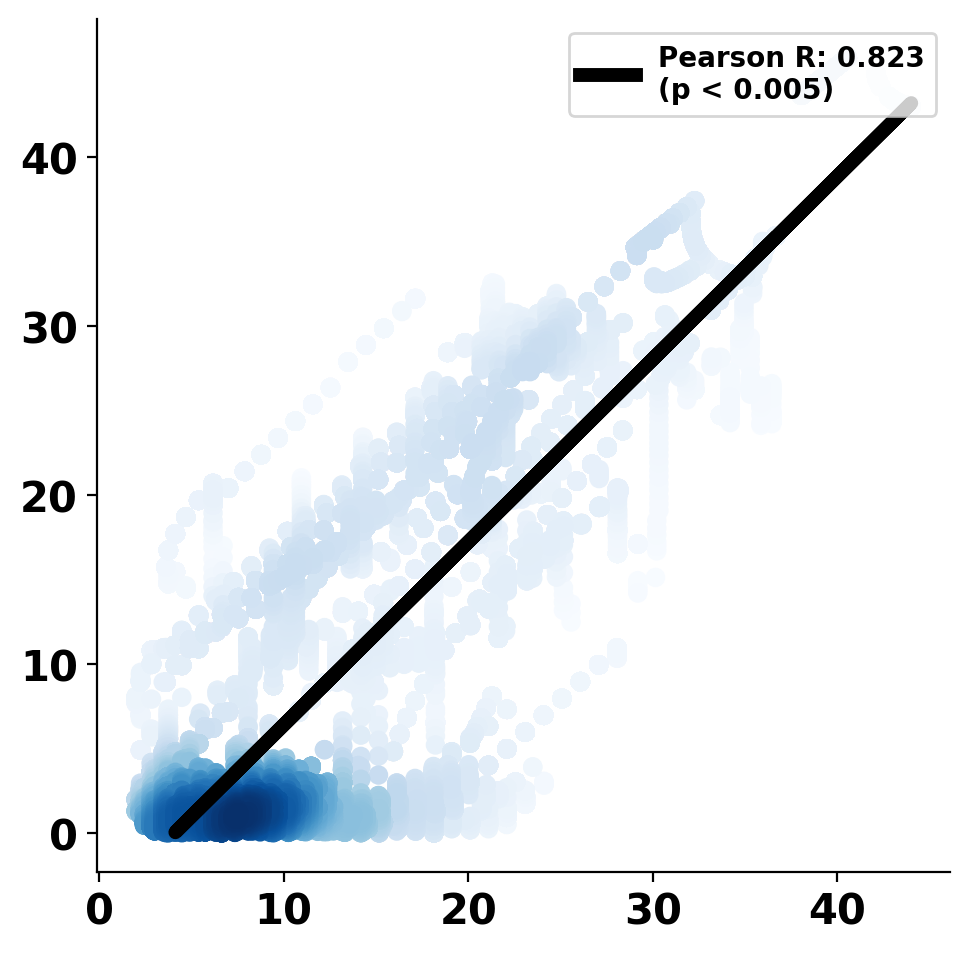} &
        \includegraphics[align=c, height=0.24\linewidth]{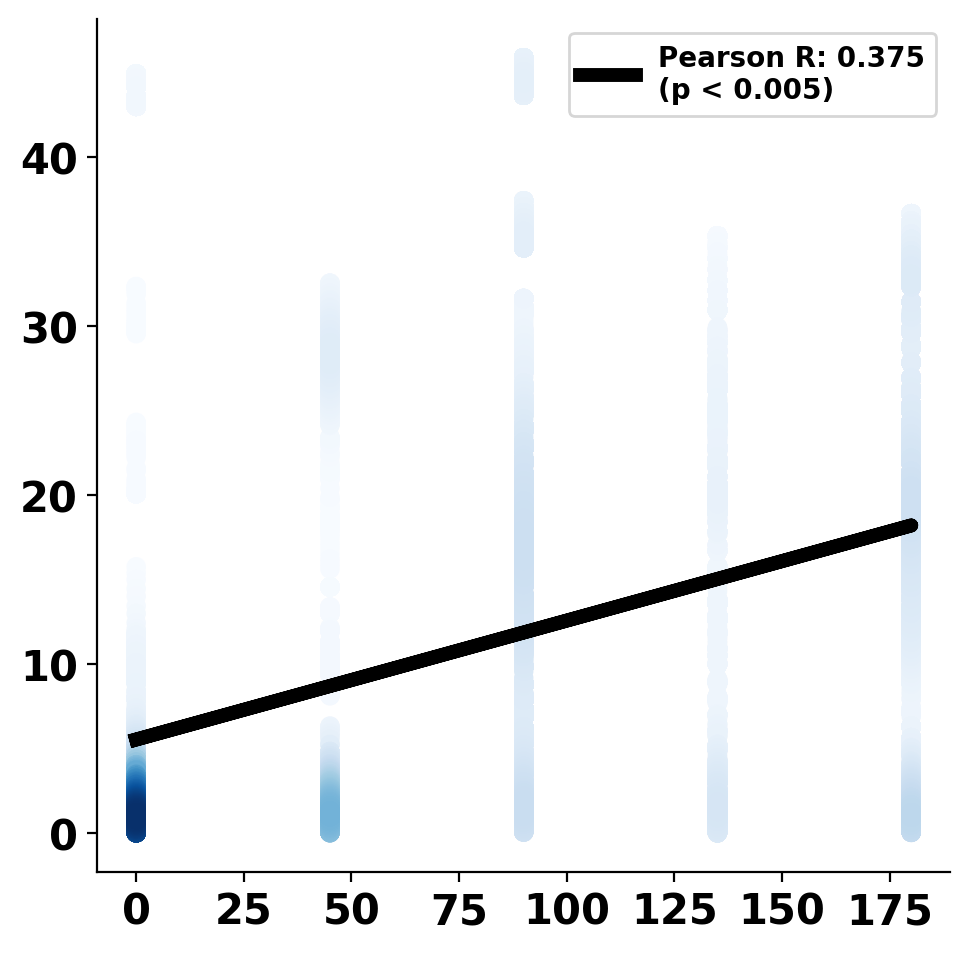} &
        \includegraphics[align=c, height=0.24\linewidth]{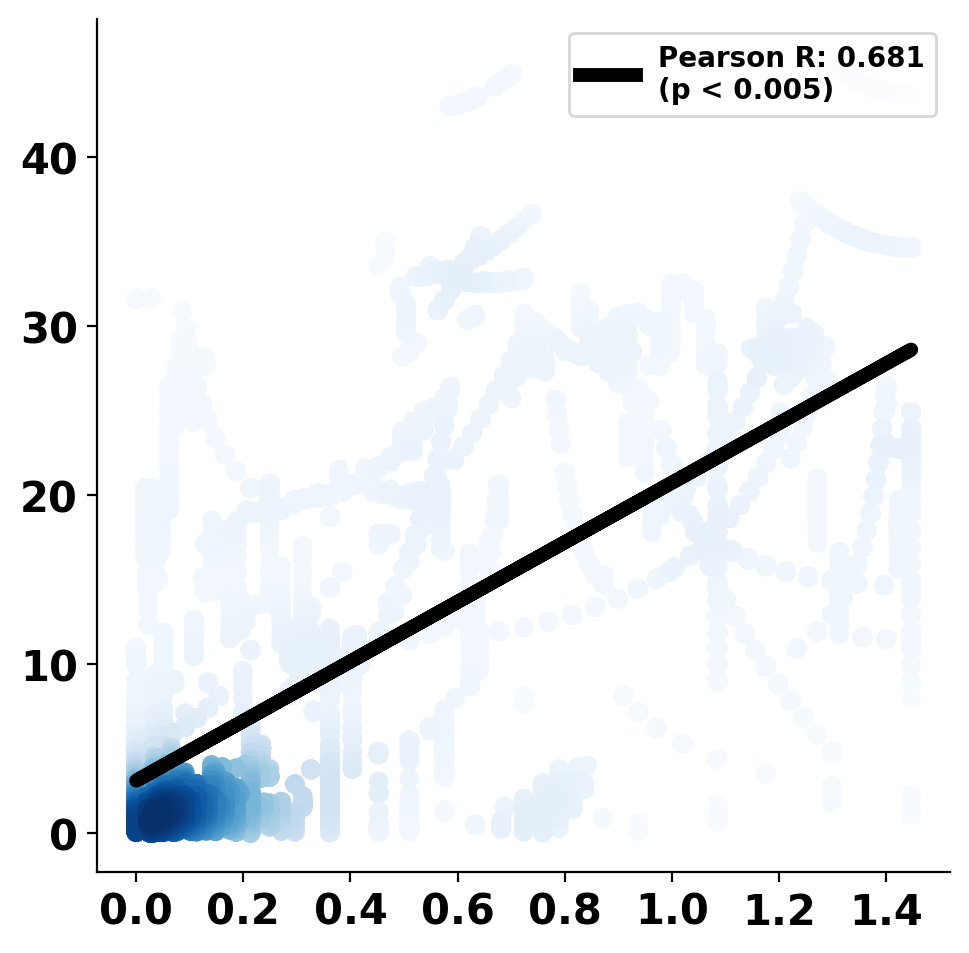} &
        \includegraphics[align=c, height=0.24\linewidth]{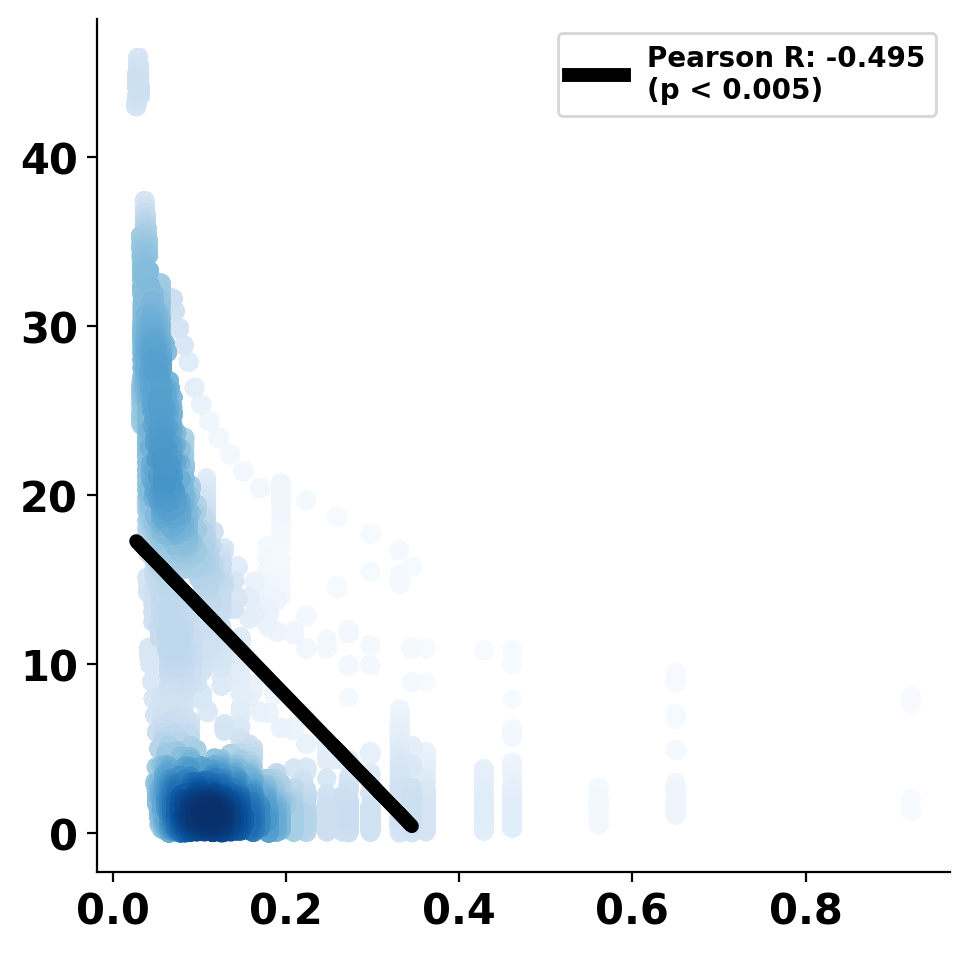} \\
        \rotatebox[origin=c]{90}{Orientation} &
        \rotatebox[origin=c]{90}{error (degree)} &
        \includegraphics[align=c, height=0.24\linewidth]{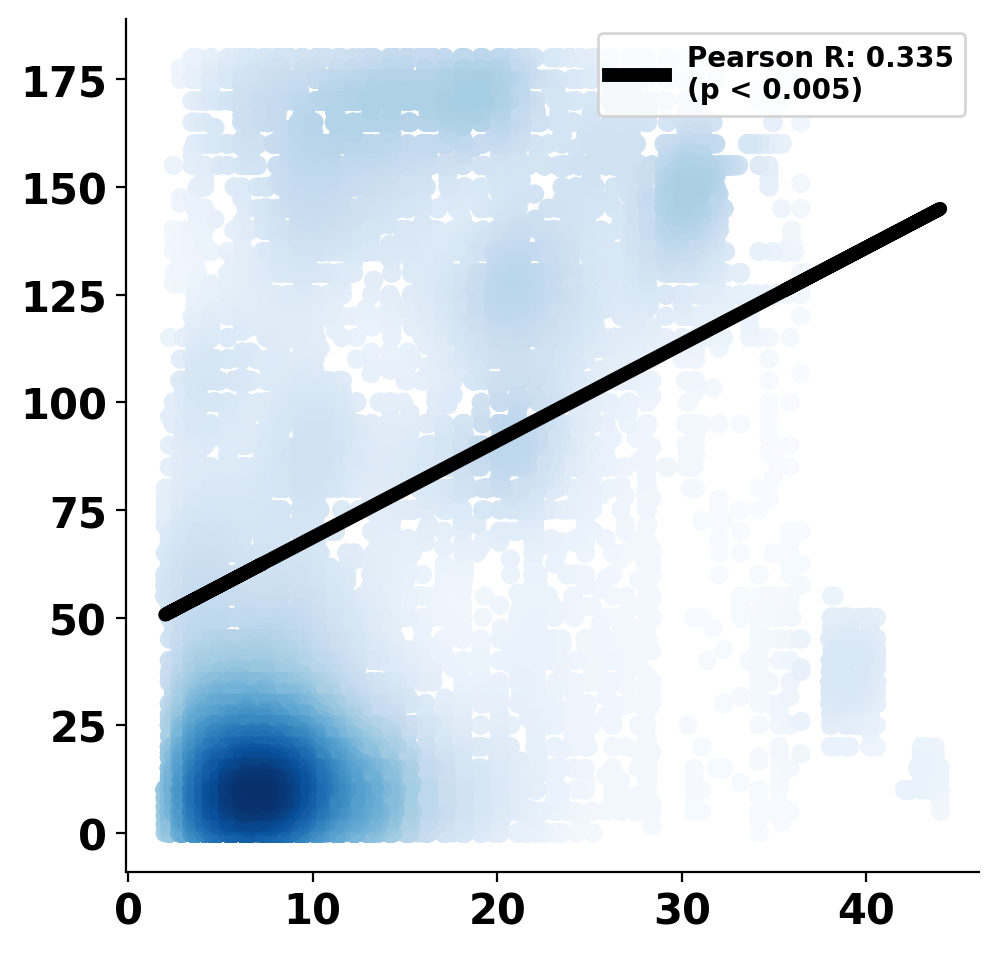} &
        \includegraphics[align=c, height=0.24\linewidth]{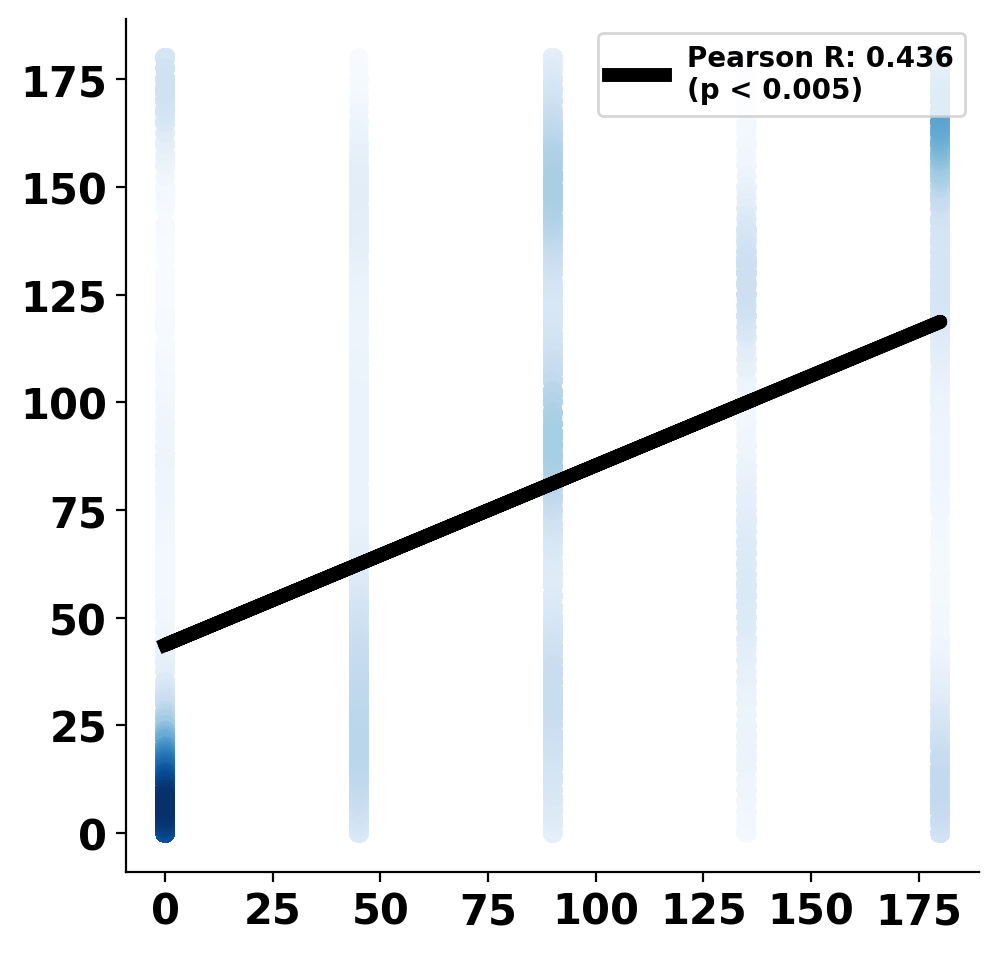} &
        \includegraphics[align=c, height=0.24\linewidth]{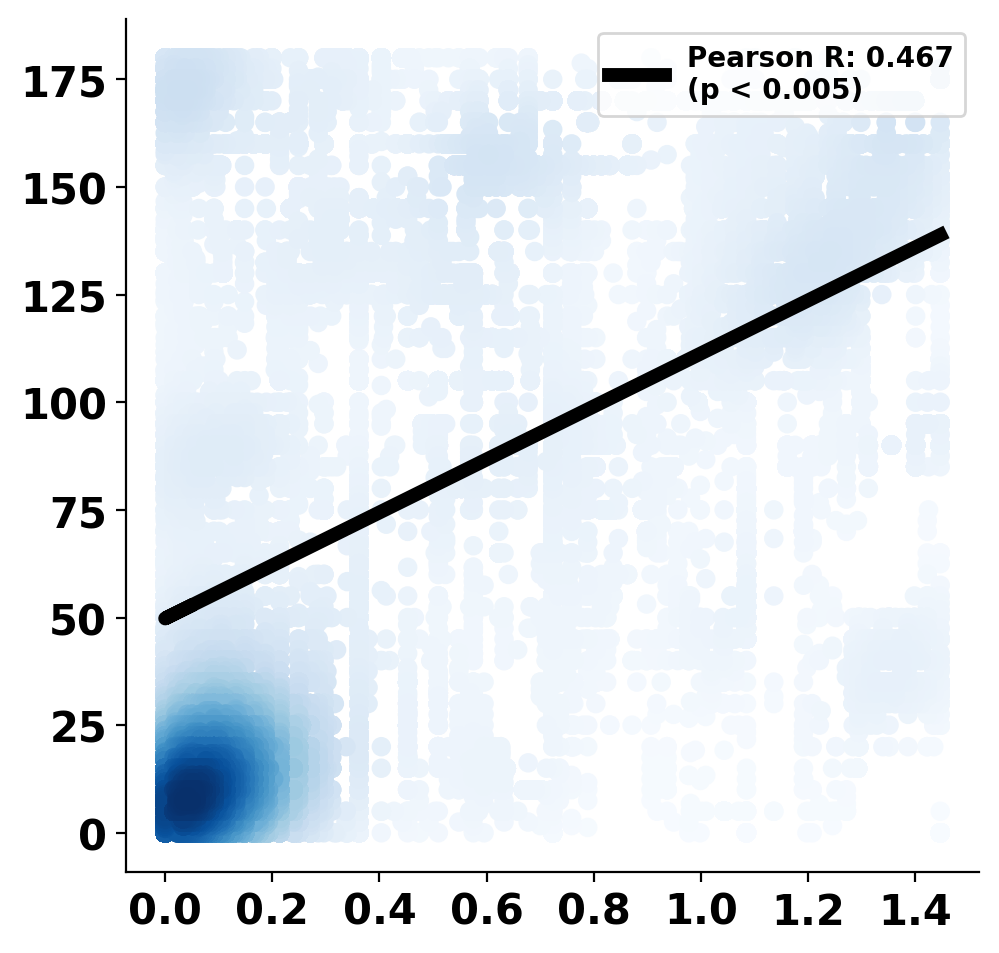} &
        \includegraphics[align=c, height=0.24\linewidth]{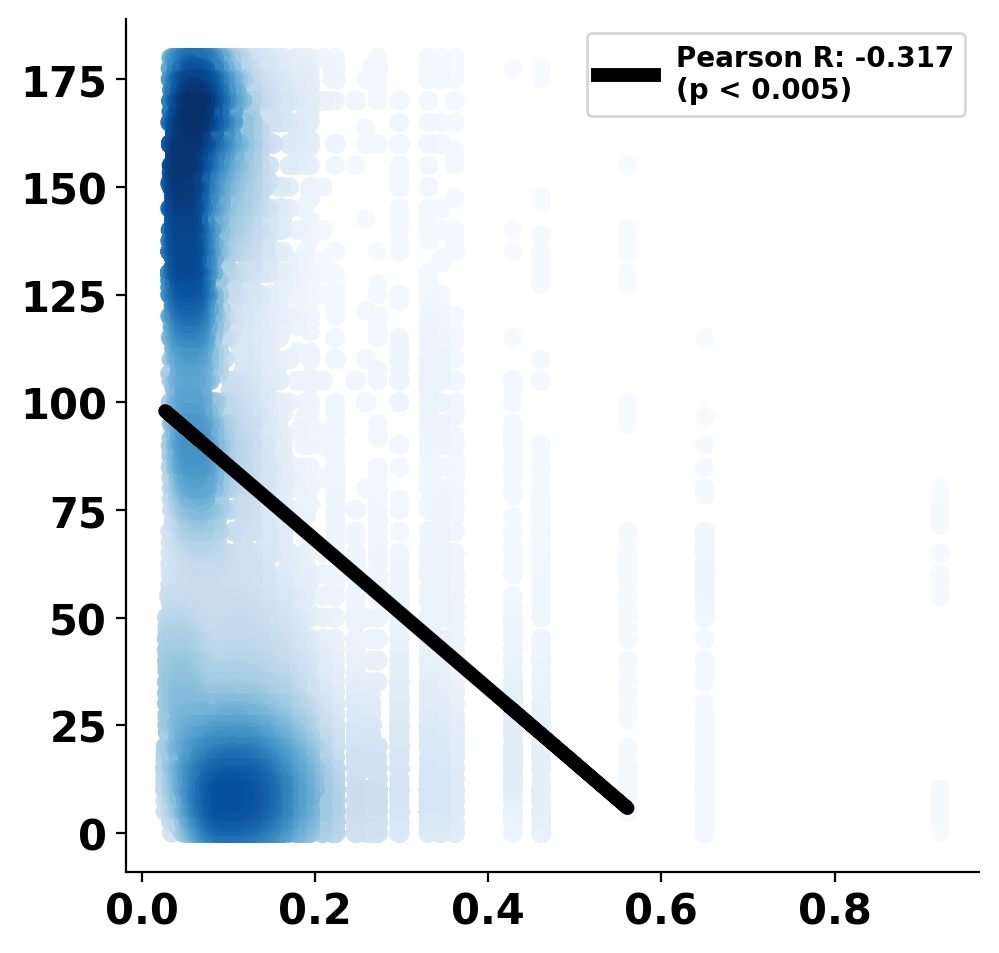} \\
        & &
        D (m) & 
        $\theta$\degree &
        $\sfrac{\theta\degree_H}{\theta\degree_{H_{FOV}}}$ &
        $\sfrac{\theta\degree_V}{\theta\degree_{V_{FOV}}}$ 
    \end{tabular}
    \end{adjustbox}
    \vspace{-0.1in}
    \caption{
    \revision{
    Localization and orientation errors regarding camera installation.
    Each of the aspects is associated with localization (1st row) and orientation estimation (2nd row) errors in each sample.
    The correlation plot shows denser samples with \textcolor{blue}{BLUE}.
    % The localization error and orientation error were strongly correlated with the distance between the camera and the captured person's location ($D$; 1st col.), the facing direction away from the camera ($\theta$; 2nd col.), and the normalized horizontal angle ($\sfrac{\theta\degree_H}{\theta\degree_{H_{FOV}}}$; 3rd col.), and they were strongly negatively correlated with the normalized vertical angle($\sfrac{\theta\degree_V}{\theta\degree_{V_{FOV}}}$; 4th col.).    
    % This shows that multiple cameras installed densely with multiple views are important, as such systems allow people to be captured from closer distances ($<15m$) and directly from their facing direction ($\theta\degree_H<30\degree$), which leads to a more accurate estimate of body orientation and localization.      
    }
    }
    \label{fig:corr_analysis}
\end{figure*}

\subsection{Impact of Parameters of Camera Installation}
\label{sec:cam_install}
In \autoref{fig:corr_analysis}, a Pearson correlation analysis is presented, highlighting the relationship between each factor from \autoref{fig:cam_install_protocol} and localization and orientation errors. In the figure, areas with denser samples are represented in a deeper shade of blue.

\subsubsection{Distance Between Camera and Person}
In \autoref{fig:corr_analysis}, in the first column, a strong positive correlation is evident between the localization error (first row) and the camera distance, especially when the distance exceeds 20 meters ($R=0.823$, $p<0.005$).
Similarly, the orientation error (first column, second row) also exhibits a positive correlation with an increase in the distance between the captured individual and the camera ($R=0.335$, $p < 0.005$).

\subsubsection{Facing Angle with Respect to Camera Viewpoint}
In \autoref{fig:corr_analysis}, the second column, a strong positive correlation is observed between the facing angle concerning the camera viewpoint and both localization ($R=0.375$, $p<0.005$) and body orientation estimation ($R=0.436$, $p<0.005$) errors.

% , and as the person turned away from the camera, more distortions were introduced to the estimated keypoint locations, which affected localization accuracy.
% This was similar to the body orientation estimation model, as the confidence of the model was high when the person was facing forward. 
% The model had more confusion as the person started facing away from the camera.

\subsubsection{Horizontal Displacement from Camera Center}
In \autoref{fig:corr_analysis}, in the third column, it is evident that the localization and body orientation estimation errors displayed significant increases, with strong positive correlations, as the captured individual moved away from the horizontal center. The correlation coefficients for localization and body orientation estimation errors were $R=0.681$ and $R=0.467$, respectively, both with $p<0.005$.

% Again, this illustrates the advantage of densely installing cameras to capture people with overlapping viewpoints of $30\degree$ horizontal angle displacements.

\subsubsection{Vertical Displacement from Camera Center}
As the individual moved closer to the vertical camera center, there was a notable reduction in both the localization and orientation estimation errors. This decrease in errors can be attributed to the increased coverage of more body parts within the camera's field of view. This relationship was marked by strong negative correlations, with correlation coefficients of $R=-0.495$ for localization and $R=-0.317$ for orientation estimation, both with $p<0.005$.

\section{Discussion}

\revision{
Our experiment results demonstrated the possibility of using edge computing devices and cameras to continuously monitor fine-grained activities across large indoor spaces.}
% In this section, we discuss the limitations and potential applications of our systems that can largely benefit from our low-cost and privacy-preserving monitoring system.

\subsection{Multi-person Tracking and Localization}

\subsubsection{4 People Walking around Entire Space}

\revision{
The evaluations shows that the proposed model consistently tracks individuals throughout the sequence as long as they are initially detected. Furthermore, the system exhibits an ability to handle a certain degree of missing detections (false negatives) arising from the limitations of the multi-person pose detection model on edge computing systems. To run the pose detection model in real-time on an edge computing system, which is a resource-constrained device, we are currently using a model that is specifically tailored to emphasize efficiency over performance compared to the state-of-the-art models~\cite{dos2021monocular}. This emphasis on efficiency is reflected in the system's high precision (93.48\%), high MT (1.83), and low ML (0.33). The overall performance of localization remains within a reasonable error range, given our tracking rate of 1 Hz, which is adequate considering that people typically walk at a rate of $1.42m/s$ in everyday activities~\cite{browning2006effects,mohler2007visual,levine1999pace}. Additionally, the precision of our ground truth annotations includes a margin of 1-meter error since participants' locations were recorded as the nearest 1-meter interval marker in the space.
}

\revision{
The model's performance exhibited variations based on the regions within the study space. In the Activity area (A), where most physical activities occur, the MOTA score was 88.63\%, with a MOTP of $1.38m$, high Precision (95.6\%) and Recall (93.62\%), which are sufficient for detecting, tracking, and understanding group activities and space usage behavior in this area. In the Activity Area (A), seven cameras were tracking movements, emphasizing the importance of multi-view cameras in noisy detection scenarios. The overlapping viewpoints of these cameras facilitated capturing people from various angles, including the front, back, and sides during walking, as demonstrated in \autoref{fig:camera_config}. 
}

\revision{
On the other hand, lower MOTA scores were observed in regions with only a few cameras covering a wide capture space. Specifically, the Kitchen, Left Corridor, and Right Corridor recorded MOTA scores of 67.16\%, 54.53\%, and 60.25\%, with corresponding MOTP values of $1.48m$, $1.92m$, and $1.75m$. These regions featured long corridors with dimensions of $24m\times 3m$, $33m\times 2m$, and $33m\times 1.5m$, respectively. The limited camera coverage for these regions, and capturing individuals from a distance, led to challenges related to occlusion and pose detection. The small pixel region allocated for each person resulted in elevated false negative rates (FNRs) of 1.01, 1.53, and 1.56, respectively. Nonetheless, our system exhibited consistently high Precision (95.6\%, 91.3\%, 100\%) in these regions, indicating that the detected objects were predominantly accurately localized.
}

\revision{
The Staff Zone is equipped with four cameras positioned to capture movements from the sides (left and right) while walking. The size of the Staff Zone, which measures $14m\times 3m$, is the smallest among all areas, leading to cameras capturing individuals at closer distances. As a result, this area achieved a higher MOTA score of 73.47\% and had a lower MOTP score of $1.11m$ and reduced false negative rates (FNRs) of 0.40 compared to other areas. Conversely, the Lounge area only features a single camera covering the entire space and exhibited} \revision{the lowest performance, with a MOTA score of 33.44\%. This was primarily attributed to the lack of power and network resources near the lounge area, limiting the use of a single camera for coverage. Additionally, occlusions from the front desk, chairs, and other objects in the lounge further impacted the tracking accuracy. In summary, our experimental results highlight the importance of multi-view cameras positioned at closer distances to individuals. These factors are essential for addressing challenges associated with noisy pose estimations in edge computing devices, ultimately leading to lower MOTP scores and improved MOTA performance.
}

\subsubsection{5 People Walking Past Each Other}
\revision{
Our findings indicate that employing four multi-view cameras to capture activities from different perspectives can help mitigate occlusions when individuals are in motion, even in the presence of noisy or missing pose estimations.
}

\subsection{Body Orientation Estimation}

\subsubsection{4 People Walking around Entire Space}

\revision{
The body-orientation estimation performance reliably distinguished between eight orientations (N, NE, E, SE, S, SW, W, NW). However, the performance of orientation estimation varied depending on the size of the captured area and the overlap between multi-view cameras. In the Activity Area, which had seven cameras capturing views from different angles, the orientation estimation errors were relatively low, with a mean absolute error (MAE) of 25.91° (\autoref{tab:orientation}, 1st row). The Staff Zone, equipped with four cameras providing overlapping views and close proximity to individuals, also exhibited lower MAE (32.41°) (\autoref{tab:orientation}, 6th row). 
}

\revision{
Surprisingly, the Left and Right Corridors, despite having low MOTA scores, demonstrated low orientation estimation errors with MAEs of 21.68° and 32.41°, respectively (\autoref{tab:orientation}, 3rd and 5th rows). These areas faced challenges in localization and tracking due to fewer cameras covering long and narrow distances. When estimating body orientations, the narrow width of the corridors made people tend to face forward or opposite to the camera viewpoints when they move, making it easier for the model to estimate body orientations.
In contrast, the Lounge area, covered by a single camera and experiencing occlusions from objects, found it challenging to identify body orientations. This resulted in a higher MAE of 47.37° and lower Accuracy-45° of 55.4\% (\autoref{tab:orientation}, 4th row). Overall, the placement of cameras is dependent on the space's structural characteristics. Fewer cameras (1 or 2) suffice for narrow corridors to capture linear movements, while open spaces benefit from multi-view cameras to account for complex and unrestricted wandering.
}

\subsubsection{5 People Walking Past Each other}

\revision{
The body orientation estimation model struggled with occlusions when people were close to each other. However, it still managed to achieve an Accuracy-90° score of 81.4\%. This indicates that our model can reliably identify the four cardinal directions (N, E, S, W) even when people are in close proximity to each other and changing group formations dynamically. These insights establish a foundation for analyzing group activities in crowded areas, as successful group activity analysis necessitates the identification of group formations and facing directions during movement.
}

\subsubsection{5 people Changing Orientations in Place}

\revision{
In this scenario, our model was capable of identifying the four cardinal directions (N, E, S, W) with a minimum accuracy of 69.3\% (Acc.-90°). Similar to the previous experiment, the primary source of error was occlusions between people who were in close proximity, resulting in multiple individuals being captured in a single human bounding box. Additionally, our body orientation estimation model struggled to detect subtle changes in facing direction (approximately 5° to 10°) between consecutive frames, as it estimates orientations frame by frame. This indicates that our model performed better when individuals were walking, as our multi-person tracking method updated the orientation information based on their movement directions.
}

\revision{
The accuracy of each session varied based on the number of cameras and the size of the areas they covered.
In Session 1 (\autoref{tab:orientation}, 9th row), which was covered by three cameras and had a space of $55m^2$, the mean absolute error (MAE) was 73.7\degree. It's worth noting that the samples in this session consisted of smaller groups with only two or three members.
Session 2 (\autoref{tab:orientation}, 10th row) had a slightly lower MAE of 65.2\degree. In this session, group sizes were larger (four members), but the covered area was smaller ($28m^2$) than in Session 1. This demonstrates the significance of multi-view cameras for improved accuracy.
Session 3 (\autoref{tab:orientation}, 11th row) was particularly noteworthy, as it had the lowest MAE (50.4\degree) among all three sessions. Despite having a larger social group of five people, the capture space was only $15m^2$, and there were four cameras available, highlighting the importance of smaller areas and more cameras for better performance.
}

\subsection{Impact of Parameters of Camera Installation}

\subsubsection{Distance Between Camera and Person}

\revision{
The majority of our samples exhibited localization errors within approximately 2 meters and orientation errors within 25 degrees when the distance between the person and the camera was less than 15 meters. This observation highlights that for the edge computing cameras, the portion of pixels corresponding to a person significantly diminishes as the person moves beyond 15 meters. The models faced more challenges when individuals obscured each other, making it appear as if multiple people were a single person when viewed from a distance. Additionally, the pose estimation models often struggled to distinguish between frontal and rear views of a person, particularly when the person was positioned over $20m$ away from the camera, resulting in distorted keypoint predictions.
}

\subsubsection{Facing Angle with Respect to Camera Viewpoint}

\revision{
The orientation at which the captured person was facing had a pronounced influence on the accuracy of both localization and body orientation estimation. Facing angles within 50 degrees of the camera's line of sight were identified as pivotal in achieving a localization error of less than 2 meters and a body orientation error of less than 25 degrees. The 2D pose estimation model exhibited its optimal performance when the person was facing the camera directly, resulting in more confident estimations.
}

\subsubsection{Horizontal Displacement from Camera Center}

\revision{
To achieve a localization error of less than $2m$ and an orientation error of less than 25 degrees, it is essential that a person is within 30\% of the horizontal field of view. When a person is captured too close to the frame boundaries, the lens distortion in the edge computing camera disrupts the planar assumption of the floor, leading to a degradation in the perspective transformation (as discussed in Sec.~\ref{sec:indoor_localization}). This disruption significantly increases the localization error. Additionally, when subjects are only partially within the frame boundaries, the orientation estimation error also escalates.
}

\subsubsection{Vertical Displacement from Camera Center}

\revision{
As illustrated in \autoref{fig:corr_analysis} (4th column), our edge computing camera module exhibited limitations when it came to capturing the foot locations of individuals with $\sfrac{\theta\degree_V}{\theta\degree_{V_{FOV}}} \geq 0.8$ normalized vertical angle. In such cases, the pose estimation model failed to detect the person. On the contrary, when the value of $\sfrac{\theta\degree_V}{\theta\degree_{V_{FOV}}}$ approached zero, both localization and body orientation errors showed a significant increase, primarily due to the greater distance between the captured person and the camera. The optimal range for the normalized vertical angle was found to be between $0.05\leq \sfrac{\theta\degree_V}{\theta\degree_{V_{FOV}}} \leq 0.2$. This corresponds to distances ranging from approximately 6 meters to 23 meters between the captured person and the camera. This observation aligns with the conclusions drawn from the analysis of horizontal camera distance discussed earlier.
}

\subsection{Limitations} 
We identified structural variances of approximately $1.5m$ compared to the building blueprint, which was used to develop our localization model in Sec.~\ref{sec:indoor_localization}. Nevertheless, we deemed a margin of 2 meters for localization error to be acceptable for effectively tracking participant behavior at our study site in future applications.

When it comes to annotating body orientation, achieving precise turns in exact intervals of $5\degree \sim 10\degree$ proved to be challenging for our participants. However, we consider an error range of up to $45\degree$ to be reasonable for tracking social interactions during group activities. The normal horizontal field-of-view of the human eye is about 135\degree, and our system can detect dyadic interactions from overlapping field-of-view between two people with at most $90\degree$ margin of error. In our analysis, the mean absolute error (MAE) for body orientation was found to be $28.97\degree$ overall, which falls well within the acceptable $45\degree$ margin of error.

The areas that were not covered in our evaluation are filled with tables, yoga mats, or gym equipment.
Future studies will examine how complex movements and occlusions affect our tracking and body orientation estimation methods when people interact with those furniture and equipment.
\revision{
Despite these limitations in our evaluation, the proposed work shows promise as an initial step towards analyzing complex movements in open spaces using a privacy-preserving camera-based activity monitoring system using secure edge computing framework.
}

\subsection{Future Applications}
\revision{
\noindent \textit{Clinical Application:}
Our system allows continuous and objective assessment of gait, spatial navigation, and group activities.
Our system can be used to understand behavior and potentially detect behavioral changes in individuals with various motor and cognitive impairments~\cite{LINDHRENGIFO202283}. 
The future work will further enable a system that may be able to predict falling episodes in real-time, thereby preventing an adverse event, by directly using 2D poses from multi-view cameras across the facility~\cite{lach2017falls,allali2017management}.
}

\revision{
\noindent \textit{Workspace Design:}
Social interactions at work, which are highly correlated with productivity and well-being, are influenced by workplace design~\cite{lee2019empirical}.
Our system can provide detailed information on socio-spatial formation patterns, such as space occupancy, behavioral mapping~\cite{cox2018understanding}, Proxemics (physical distance between the people)~\cite{ballendat2010proxemic,hall1963system}, F-formation (shape and size of group formation)~\cite{danninger2005using,kendon1990spatial}, which allows the designer to take into account to design the workplace.
}

\section{Conclusion}
\revision{
Camera-based movement monitoring systems are gaining popularity thanks to the progress in edge computing platforms and computer vision technologies. In this work, we demonstrated the practicality of employing edge computing cameras to localize, estimate body orientations and track multiple individuals within a large indoor space while not compromising on privacy. Our sensing hub leveraged cost-effective edge computing components, including a Raspberry Pi, its camera module, a Tensor Processing Unit (TPU), and a computing infrastructure that can be easily installed in pre-existing building setups.
}

In our study, the described system demonstrated highly reliable multi-person tracking ($62.0\sim88.6\%$ MOTA) and body orientation estimation (29.0\degree) across various functional spaces.
For the region with more cameras ($7>$), our system could track people in crowded activities with a higher level of accuracy (88.6\% MOTA).
Additionally, we anticipate that our comprehensive examination of camera installation will offer valuable insights for individuals interested in implementing our system to enhance their environments. In our future work, we plan to increase the camera density based on the insights from our analysis. We will also assess the viability of tracking social interactions in more intricate settings, such as those involving group activities amid furnishings or the use of gym equipment.

\section{References}

\bibliographystyle{IEEEtran}
\bibliography{
bibs/orientation,
bibs/localization,
bibs/tracking,
bibs/vision,
bibs/apps,
bibs/interaction,
bibs/privacy,
bibs/edge,
bibs/mci,
bibs/design}

\end{document}